\documentclass[lettersize,journal]{IEEEtran}
\usepackage{amsmath,amsfonts}
\usepackage{algorithmic}
\usepackage{algorithm}
\usepackage{array}
\usepackage{siunitx}
\usepackage[caption=false,font=normalsize,labelfont=sf,textfont=sf]{subfig}
\usepackage{textcomp}
\usepackage{stfloats}
\usepackage{url}
\usepackage{verbatim}
\usepackage{graphicx}
\usepackage{cite}
\usepackage{comment}
\usepackage{amssymb}
\usepackage{hyperref}

\newtheorem{theorem}{Theorem}

\begin{document}

\title{Neural Koopman prior for data assimilation}

\author{Anthony Frion,~\IEEEmembership{Student Member,~IEEE}, Lucas Drumetz,~\IEEEmembership{Member,~IEEE,} Mauro Dalla Mura,~\IEEEmembership{Senior Member,~IEEE,} Guillaume Tochon, Abdeldjalil Aïssa El Bey
\thanks{\noindent A. Frion, L. Drumetz and A. Aïssa El Bey are with IMT Atlantique, Lab-STICC, UMR CNRS 6285, Brest, France.\\
M. Dalla Mura is with Univ. Grenoble Alpes, CNRS, Grenoble INP, GIPSA-lab, Grenoble, France, and Institut Universitaire de France.\\
G. Tochon is with LRE Epita, Le Kremlin-Bicêtre, France.\\
This work was supported by Agence Nationale de la Recherche under grant ANR-21-CE48-0005 LEMONADE.}}%

\markboth{Transactions on Signal Processing}%
{Shell \MakeLowercase{\textit{et al.}}: A Sample Article Using IEEEtran.cls for IEEE Journals}


\maketitle

\begin{abstract}
With the increasing availability of large scale datasets, computational power and tools like automatic differentiation and expressive neural network architectures, sequential data are now often treated in a data-driven way, with a dynamical model trained from the observation data. While neural networks are often seen as uninterpretable black-box architectures, they can still benefit from physical priors on the data and from mathematical knowledge. In this paper, we use a neural network architecture which leverages the long-known Koopman operator theory to embed dynamical systems in latent spaces where their dynamics can be described linearly, enabling a number of appealing features. We introduce methods that enable to train such a model for long-term continuous reconstruction, even in difficult contexts where the data comes in irregularly-sampled time series. The potential for self-supervised learning is also demonstrated, as we show the promising use of trained dynamical models as priors for variational data assimilation techniques, with applications to e.g. time series interpolation and forecasting. 
\end{abstract}


\section{Introduction}
\label{Introduction}

The evergrowing amount of historical data for scientific applications has recently enabled to model the evolution of dynamical systems in a purely data-driven way
using powerful regressors such as
neural networks. While many of the most spectacular results obtained by neural networks rely on the paradigm of supervised learning, this paradigm is limited in practice by the available amount of labelled data, which can be prohibitively costly and difficult to obtain. For example, a number of Earth observation programs that have been launched in the last decade provide huge amounts of sequential (though generally incomplete) satellite multi/hyperspectral images covering the entire Earth's surface. However, few accurate and reliable labels exist for land cover classification of the ground pixels, although some efforts have been made, e.g. for crop type classification and segmentation~\cite{BreizhCrops, TimeSen2Crop}. 

In this context, one can leverage another machine learning paradigm called self-supervised learning (SSL)~\cite{SSL_review}. It consists in training a machine learning model to solve a pretext task that requires no labels in order to learn informative representations of the data which can be used to solve downstream tasks.
When dealing with image data, possible pretext tasks include predicting the relative positions of two randomly selected patches of a same image~\cite{SSL1} and predicting which rotation angle has been applied to an image~\cite{SSL2}. Many SSL approaches can be labelled as contrastive SSL~\cite{SSL3}, which means that they aim at learning similar representations for images that are related by transformations such as rotations, crops and color transfers. We refer the interested reader to~\cite{SSL-RM} for a review of self-supervised learning for remote sensing applications.

In our case, since we are dealing with sequential data, we use a natural pretext task which consists in being able to forecast the future state of the data from a given initial condition. This is similar in spirit to recent approaches in natural language processing where a model is trained on completing texts and then used to perform various tasks in zero/few shot, e.g.~\cite{GPT3}. Our trained model is used for downstream tasks which can be formulated as inverse problems such as denoising and interpolation. We solve these tasks by minimising a variational cost which uses a trained model as a dynamical prior, contrarily to classical data assimilation techniques~\cite{Evensen,borsoi2020kalman, borsoi2023dynamical} which leverage hand-crafted dynamical priors that require domain knowledge, and are not always available. Besides, these priors should be differentiable since any first order optimisation algorithm tackling such a problem must be able to differentiate through repeated compositions of the model, which requires careful implementation~\cite{lewis1985use} and is out of reach for many operational systems relying on complex dynamical models~\cite{nonnenmacher2021deep}. In contrast, neural emulators of the dynamics are \emph{de facto} implemented in packages supporting automatic differentiation, e.g. Pytorch~\cite{paszke2017automatic}, Tensorflow, JAX, etc. providing effortless access to model derivatives.

For all these reasons, in this paper, we first aim at modelling dynamical systems from observation data using differentiable models. We assume that the state of a dynamical system can be described by a $n$-dimensional state variable $\mathbf{x} \in \mathcal{D}$ with $\mathcal{D} \subset \mathbb{R}^n$. Then, assuming that the system is governed by an autonomous ordinary differential equation (ODE), one can describe its (discrete) dynamics by a function $F : \mathcal{D} \to \mathcal{D}$ such that $\mathbf{x}_{t+1} = F(\mathbf{x}_t)$. Although $F$ might be any non-linear function, Koopman operator theory~\cite{Koopman} tells us that the system can be described by a linear operator acting in the space of observation functions. Namely, given an observation function $f : \mathcal{D} \to \mathbb{R}$, the so-called Koopman operator $\mathcal{K}$ composes $f$ by a time increment through function $F$:
\begin{equation}
    \mathcal{K}f(\mathbf{x}_t) \triangleq (f \circ F) (\mathbf{x}_t) = f(\mathbf{x}_{t+1}).
\end{equation} 
From this definition, $\mathcal{K}$ is linear because of the linearity of the function space, i.e. for any $f,g : \mathcal{D} \to \mathbb{R}$:
\begin{equation}
    \mathcal{K}(f + g) (\mathbf{x}_t) = (f+g)(\mathbf{x}_{t+1}) = \mathcal{K}f(\mathbf{x}_t) + \mathcal{K}g(\mathbf{x}_t).
\end{equation} 
Yet, the function space being infinite dimensional, the advantage of the linearity of $\mathcal{K}$ comes at the cost of an infinite dimension, which makes it difficult to model in practice. 
Thus, for practical purposes, it is particularly interesting to study finite sets of linearly independent observation functions $(\phi_1,...,\phi_d)$ from which the span is invariant by the Koopman operator: such a span is called a Koopman invariant subspace (KIS)~\cite{brunton2021modern}. A function $f$ from this space may be written in its basis with a vector $\mathbf{a} \in \mathbb{R}^d$ such that 
\begin{equation}
\label{eq_decomposition}
    f(\mathbf{x}_t) = \sum_{1 \leq i \leq d} a_i \phi_i(\mathbf{x}_t).
\end{equation} 
Then, the action of the Koopman operator on $f$ can be summarized in the same basis by a vector $\mathbf{b} \in \mathbb{R}^d$ such that
\begin{equation}
\label{eq_restriction}
    \mathcal{K}f(\mathbf{x}_t) = f(\mathbf{x}_{t+1}) = \sum_{1 \leq i \leq d} b_i \phi_i(\mathbf{x}_t), \quad \mathbf{b} = \mathbf{K}\mathbf{a},
\end{equation}
where $\mathbf{K} \in \mathbb{R}^{d \times d}$ is the $d$-dimensional restriction of the infinite-dimensional $\mathcal{K}$ on the KIS spanned by $(\phi_1,...,\phi_d)$. Writing such a matrix is only possible because the span of $(\phi_1,...,\phi_d)$ is a KIS, otherwise $\mathcal{K}f$ might not be included in this span.

There exists a variety of KIS, but one needs to retrieve nontrivial ones that give information about the dynamics of the state variable. 
As a basic example, consider a discrete $n$-dimensional linear dynamical system, described by a state variable $\mathbf{x}_t = ({x}_{t,1},...,x_{t,n})^\intercal \in \mathbb{R}^n$, and which evolves in time through 
\begin{equation}
\label{eqLinearDS}
    \mathbf{x}_{t+1} = F(\mathbf{x}_t) = \mathbf{A}\mathbf{x}_t,
\end{equation} with $\mathbf{A} \in \mathbb{R}^{n \times n}$. In this case, the state observation functions defined by $f_i(\mathbf{x}_t) = x_{t,i}$ for $1 \leq i \leq n$ form a KIS, and the associated restriction of $\mathcal{K}$ is simply $\mathbf{K} = \mathbf{A}$. Moreover, some nonlinear dynamical systems feature a finite-dimensional KIS which includes the state observation functions, allowing to reformulate them exactly as higher-dimensional linear dynamical systems: see section 5.1 of~\cite{bruntonKIS} for an example.

Once a KIS is found, the associated $\mathbf{K}$ can be interpreted with classical linear algebra tools. Notably, each of the complex eigenvalues of $\mathbf{K}$ is associated to an observation function that is located in the subspace. Let us denote by $\mathbf{K} = \mathbf{V}\mathbf{\Lambda}\mathbf{V}^{-1}$ the complex eigendecomposition of $\mathbf{K}$, with $\mathbf{V}$ the complex eigenvectors and $\mathbf{\Lambda}$ a complex diagonal matrix containing the associated eigenvalues. Predicting $\tau$ steps in the future through the Koopman operator means multiplying the initial latent state vector (obtained with the functions from the invariant subspace) by $\mathbf{K}^\tau = \mathbf{V}\mathbf{\Lambda}^\tau\mathbf{V}^{-1}$. Therefore, the eigenvectors associated with an eigenvalue of modulus higher than one will have an exponentially growing contribution, while those with an eigenvalue of modulus smaller than one will exponentially vanish. Only eigenvalues of modulus very close to one will approximately preserve the norm of the latent state in the long run, which might be crucial for time series with clear seasonality or periodicity.

Our approach fits into the Koopman operator framework to model dynamical systems from data. More specifically, our contributions are the following: 

1) We perform a synthetic review of the different approaches that have recently been used to compute data-driven approximations of the Koopman operator, emphasising on the limitations of each of the successive categories of approaches.

2) We refine and extend our own approach to learn a neural Koopman operator, first sketched in~\cite{ICASSP}, with theoretical arguments to support the search for an approximately orthogonal Koopman operator and extensive evidence that this search is beneficial for training models that transfer well to new tasks and data distributions.

3) We present in detail the conditions and methods for switching our model from a discrete to a continuous formulation and vice versa, in order to train and evaluate on irregularly-sampled sequential data. To the best of our knowledge, we present the first experiments of training a data-driven Koopman model on irregular data.

4) We discuss several ways to use our model as a fully-differentiable dynamical prior in data assimilation in order to solve inverse problems using automatic differentiation. We present extensive experiments for forecasting and interpolation, including in hard scenarios such as irregularly sampled data and transfer to areas unseen during training. We show that our model is a stronger and more versatile choice as a learned dynamical prior than classical forecasting models such as long short-term memory (LSTM) neural networks.



\section{Background and related works}
\label{RelatedWorks}

\subsection{Koopman operator theory}

In short, the Koopman operator theory~\cite{Koopman} states that any dynamical system can be described linearly at the cost of an infinite dimension.
However, some methods seek to find a finite-dimensional representation of the Koopman operator. 
Such a representation can be exact only if it is associated to a Koopman invariant subspace (KIS) of the dynamical system.

Let us suppose again that we are working with a state variable $\mathbf{x} \in \mathcal{D}$ with $\mathcal{D} \subset \mathbb{R}^n$, and that the state observation functions $f_i$ are defined as
\begin{equation}
    \mathbf{x} = (f_1(\mathbf{x}), f_2(\mathbf{x}), ..., f_n(\mathbf{x}))^\intercal.
\end{equation}
Dynamic Mode Decomposition~\cite{DMD} (DMD) consists in finding a matrix $\mathbf{A} \in \mathbb{R}^{n \times n}$ such that the residual $\mathbf{r}_t$ in
\begin{equation}
\label{eqDMD}
    \mathbf{x}_{t+1} = \mathbf{A}\mathbf{x}_t + \mathbf{r}_t
\end{equation}
is as small as possible in the least squares sense. This approach has been theoretically linked to the Koopman mode decomposition in~\cite{DMD-KMD}, and has known many different variants, e.g.~\cite{OptimizedDMD, ExactDMD, MultiresDMD}. However, it relies on the implicit assumption that $(f_1,...,f_n)$ spans a KIS. Indeed, having no residual error in equation~\eqref{eqDMD} would bring us to the previously examined case of~\eqref{eqLinearDS}.
This assumption can be useful in regions of the state space where the dynamics are close to linear, but it is very unlikely to be generally true. In order to mitigate this shortcoming, the Extended Dynamic Mode Decomposition~\cite{EDMD} (EDMD) uses a manually designed dictionary of observation functions $\Phi$ from the dynamical system. Common choices of dictionaries include polynomials of the observed variables up to a given degree and sets of radial basis functions. These dictionaries all include the identity of the state space, so that they can be expressed as
\begin{equation}
    \Phi(\mathbf{x}) = (\phi_1(\mathbf{x}), \phi_2(\mathbf{x}), ..., \phi_d(\mathbf{x}))
\end{equation}
where, for $1 \leq i \leq n, \phi_i(\mathbf{x}) = f_i(\mathbf{x})$, necessarily $d \geq n$, and the special case $d=n$ is equivalent to a classical DMD. The idea of the method is to design a dictionary $\Phi$ which is approximately invariant by the Koopman operator. 
One can then compute the matrix $\mathbf{K}$ that minimizes the residual $\mathbf{r}_t$ in
\begin{equation}
\label{eq:EDMD}
    \Phi(\mathbf{x}_{t+1}) = \mathbf{K}\Phi(\mathbf{x}_t) + \mathbf{r}_t
\end{equation}
in the least-squares sense. Again, the residual $\mathbf{r}_t$ can be reduced to zero for any $\mathbf{x}_t$ if and only if $(\phi_1,...,\phi_d)$ spans a KIS of the dynamical system. Since the state observation functions $f_i$ are included in $\Phi$, at inference time one can trivially project $\mathbf{K}\Phi(\mathbf{x}_t)$ onto its first $n$ components in order to get an estimation for $\mathbf{x}_{t+1}$ from $\mathbf{x}_t$.
Alternatively, gEDMD~\cite{klus2020data} approximates the continuous rather than a discrete formulation of the Koopman operator, also using a hand-designed dictionary of observation functions.
Being a generalisation of DMD, EDMD can give satisfactory results when the chosen dictionary of functions is well suited to the considered dynamical system.

However, a hand-designed dictionary of observables might still not be the most optimal choice, and it is typically very high dimensional. For these reasons, subsequent works have focused on finding methods to obtain lower dimensional dictionaries of observables.
For example, there is a rich literature on leveraging Reproducing Kernel Hilbert Spaces to obtain approximations of the Koopman operator with some interpretability and theoretical guarantees, e.g.~\cite{RKHS16, RKHS22, RKHS23}.
Recently, \cite{kostic2023learning} proposed a method for learning a representation of a dynamical system that can be used in the framework of operator regression~\cite{RKHS23} for solving downstream tasks.

Other methods~\cite{EDMD-DL, Yeung} jointly learn the parameters of a neural network which computes a set of observation functions and a matrix $\mathbf{K}$ which is the restriction of the Koopman operator to this set. 
Therefore, the main difference between EDMD and these works is that the dictionary $\Phi$ is automatically learned by a neural network rather than hand-designed.
However, to be able to retrieve the evolution of the state variable from the KIS, they still constrain that $\phi_i = f_i$ for $1 \leq i \leq n$.
This is a convenient trick, yet it restricts those methods since it means assuming that there exists a low-dimensional KIS containing the state functions.

In order not to rely on this assumption anymore, some other works~\cite{Lusch, LRAN, Morton, CompositionalKoopman, Azencot, ICASSP} do not constrain a trivial link between the KIS and the state space spanned by $(f_1,...,f_n)$. In this case, it is necessary to train another neural network to reconstruct the state variables from the learnt observation functions. Then, the network learning the KIS and the network that reconstructs the state space from it form an autoencoder by definition. This framework is theoretically more powerful than the previous ones since it only assumes a nonlinear relationship between the KIS and the state space. The model that we describe in Section~\ref{Training} and use throughout this work belongs to this category of approaches.

Of particular interest for this study are~\cite{Lusch} and~\cite{Azencot}, which we will compare our model against in Section~\ref{ExperimentsSimulated}. The model of~\cite{Lusch}, which we refer to as DeepKoopman, learns an auxiliary network that outputs eigenvalues as a function of the encoded state of the system. Therefore, it computes a new matrix $\mathbf{K}$ for each time step, while most Koopman autoencoder models, including ours, learn a fixed matrix $\mathbf{K}$ which is invariantly used for advancing any encoded state in time. The model of~\cite{Azencot} learns two distinct matrices $\mathbf{K} \in \mathbb{R}^{d \times d}, \mathbf{D} \in \mathbb{R}^{d \times d}$ for the forward and backward evolutions. They are respectively trained with a forward and a backward prediction loss, and linked through an additional consistency loss term in order to favor the consistency of the latent dynamics. This consistency loss is expressed as
\begin{equation}
\label{eq:consistencyLoss}
    \sum_{k=1}^d \frac{1}{2k}||\mathbf{K}_{k*}\mathbf{D}_{*k} - \mathbf{I}_k||_F^2 + \frac{1}{2k}||\mathbf{D}_{k*}\mathbf{K}_{*k} - \mathbf{I}_k||_F^2,
\end{equation}
where $\mathbf{M}_{k*}$ and $\mathbf{M}_{*k}$ respectively denote the upper $k$ rows and leftmost $k$ columns of a matrix $\mathbf{M}$, and $||.||_F$ is the Frobenius norm. This loss term roughly means that $\mathbf{K}$ and $\mathbf{D}$ are encouraged to be close to each other's inverse. 
The computational cost of computing it scales in $\mathcal{O}(d^4)$, so that it can only be used with a reasonably small latent dimension $d$.

For the sake of completeness, we mention works that have focused on finding matrices $\mathbf{K}$ that verify some analytical properties enabling them to produce asymptotically stable latent dynamics. The authors of~\cite{erichson2019physics} introduced a regularisation term that promotes the asymptotic stability of the learnt model in the sense of Lyapunov. In \cite{pan2020physics}, a parameterisation of stable operators using tridiagonal matrices is proposed. The authors of~\cite{fan2022learning} leverage a parameterisation of Schur stable matrices to learn a stable model. The work of~\cite{bevanda2022diffeomorphically} proposes to restrict the search to the set of Hurwitz matrices, through a parameterisation of this set that enables to solve an unconstrained optimisation problem. Most closely to our work, the Hamiltonian neural Koopman operator~\cite{zhang2024learning} consists in learning an exactly orthogonal Koopman matrix by leveraging a parameterization of orthogonal matrices as matrix exponentials of antisymmetric matrices. Overall, the methods for learning a stable Koopman model can be separated in two categories: those which use a hard constraint to ensure that the learnt matrix belongs to a chosen set~\cite{pan2020physics, fan2022learning, bevanda2022diffeomorphically, zhang2024learning} and those which use a soft penalty term~\cite{Azencot, erichson2019physics}. The method that we introduce in section~\ref{Training} belongs to the second category.

\subsection{Orthogonality regularisation}

The promotion of orthogonality for the weight matrices of linear layers in neural networks has been long studied. 
This idea is related to the well-known vanishing gradient and exploding gradient issues. Those get more important as the computational graph gets deeper, e.g. for recurrent neural networks and for very deep residual neural networks~\cite{ResNet}.

In \cite{ExactSolutions}, it was shown that the initialisation of weights as a random orthogonal matrix can be much more effective than the classical random Gaussian initialisation. 
It was also advocated that the orthogonality of the weight matrices should be promoted  during the training phase too. The authors of~\cite{BeyondGoodInit} introduced a soft regularisation term for weight matrices $\mathbf{W}$:
\begin{equation}
    ||\mathbf{W}\mathbf{W}^T - \mathbf{I}||_F^2
    \label{ortho}
\end{equation}
where $||.||_F$ is the Frobenius norm. This term, which is to be used in a similar way to weight decay~\cite{WeightDecay}, was shown to improve the performance of neural networks for computer vision tasks. In~\cite{GainMore}, it was compared with similar orthogonality-promoting terms, and showed that they all brought substantial gains to the performance of deep residual networks. 

In our case, constraining the Koopman operator to be orthogonal leads to periodic dynamics (see theorem~\ref{thm:periodic}), which are of course stable in the long run and useful to model seasonality in time series. Yet, working with an exactly orthogonal $\mathbf{K}$ may not always be desirable, for instance when the data are noisy, or the time series is not exactly periodic (e.g. when there are interannual variations or slower trends in seasonal dynamics).
For these reasons, we will resort to a soft penalisation as in~\eqref{ortho} instead of enforcing exactly the orthogonality of $\mathbf{K}$.

\subsection{Variational data assimilation}


Data assimilation consists in combining a numerical model of a physical system with a set of partial observations in order to infer the full state of the system, accounting for both uncertainties in the model and in the data. It has been used in a variety of fields, including numerical weather prediction~\cite{NWPDA}, crop yield estimation~\cite{CropDA}, surface water quality modeling~\cite{SurfaceWaterQualityDA} and forest inventory~\cite{ForestInventoryDA}. In addition to sequential data assimilation, represented by Kalman filtering and its extensions, another frequently used framework is variational data assimilation, on which we focus in this work. Variational assimilation techniques consist in minimising a differentiable cost function in order to fit a set of observations while following dynamics specified by a physical model. This classically includes backpropagating through the physical model, which has been extensively researched in so called \emph{adjoint} methods~\cite{bannister2017review}. This has recently been made easier by modern automatic differentiation packages, so long as the physical model is differentiable and can be easily reimplemented in those packages (which is not the case of all operational physical models).

In the case of a discrete dynamical system, one seeks to approximate an unknown trajectory $\mathbf{x}^{true} = (\mathbf{x}_0^{true}, \mathbf{x}_1^{true}, ..., \mathbf{x}_T^{true})$.
To do so, one leverages a noisy and/or partial set of observations $\Tilde{\mathbf{x}} = (\Tilde{\mathbf{x}}_t)_{t\in H}$ where $H \subset [\![0,T ]\!]$ is the set of times on which $\Tilde{\mathbf{x}}$ is defined. In a remote sensing context, the reason for the existence of missing time indexes might be that some observations are removed because they are obviously corrupted, for example when a cloud blocks or obfuscates the surface for a satellite sensor. Furthermore, for indexes $t \in H$, $\Tilde{\mathbf{x}}_t$ is often a noisy observation of $\mathbf{x}_t^{true}$, which might be due to the imperfection of the sensor or to a variety of physical phenomena. Making assumptions on the probability distribution of $\mathbf{x}^{true}$ can help to design a prior $\mathcal{R}$ on the desired solution.
Using $\mathcal{R}$ as a regularisation, one then seeks to find a trajectory $\mathbf{x} = (\mathbf{x}_0,...,\mathbf{x}_T)$ which is a good compromise between the fidelity of $\mathbf{x}$ to $\mathbf{\Tilde{\mathbf{x}}}$ and the prior $\mathcal{R}(\mathbf{x})$. This is done by minimising the cost function

\begin{equation}
\label{GeneralDAEquation}
    \mathcal{C}(\mathbf{x}) = \mathcal{D}(\mathbf{x},\Tilde{\mathbf{x}}) + \mathcal{R}(\mathbf{x})
\end{equation}
where $\mathcal{D}$ is a chosen discrepancy, such as a norm of the difference between two elements (restricted to times $t \in H$). Common examples of prior terms $\mathcal{R}(\mathbf{x})$ include smoothness priors, which favor the proximity between consecutive values of $\mathbf{x}$.
Most of these priors can be interpreted in a bayesian sense~\cite{wikle2007bayesian}.
In practice, when all terms vary smoothly, the cost can be minimised by gradient descent or related first order algorithms. The gradient can be obtained analytically when tractable, or using automatic differentiation, as made easily accessible by modern computing frameworks, e.g. Pytorch.

\begin{figure}
    \centering
    \includegraphics[width=8.5cm]{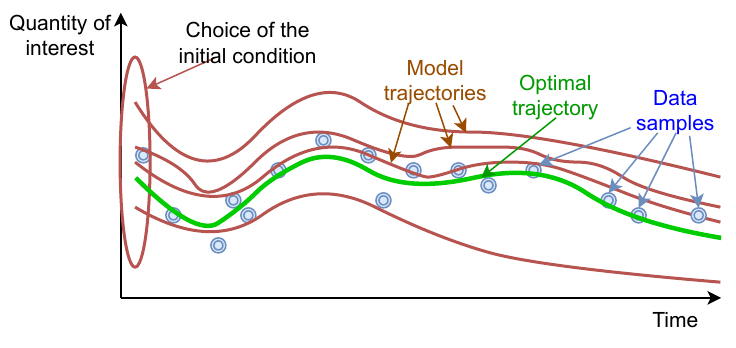}
    \caption{Visual representation of constrained variational data assimilation. It consists in choosing the initial condition from which the model's trajectory minimises the distance to the sampled data. One could also include a prior in the variational cost on the initial condition, such as the trajectory smoothness.}
    \label{fig:VDA}
\end{figure}

Alternatively, one can restrain the search on a set of trajectories defined by a model $\mathcal{M}: \mathbf{x}_0\to \mathbf{x}$. In this case, one formulates a cost on the input of the model:
\begin{equation}
    \mathcal{C}(\mathbf{x}_0) = \mathcal{D}(\mathcal{M}(\mathbf{x}_0), \Tilde{\mathbf{x}}) + \mathcal{R}(\mathcal{M}(\mathbf{x}_0)).
\end{equation}
A conceptual view of this method, called constrained variational data assimilation, is shown on Figure~\ref{fig:VDA}. We refer the reader to~\cite{Evensen} for an extensive review on data assimilation.

While variational data assimilation is traditionally used with priors $\mathcal{R}$ that were constructed from physical knowledge of the studied dynamical system, recent works~\cite{nonnenmacher2021deep, 4DVarNet} have leveraged machine learning tools to learn a prior in a completely data-driven way. In the second case, the prior is jointly learned with a gradient-based optimisation algorithm, further improving the performance. Other works~\cite{MLCorrect} have proposed to learn a data-driven surrogate model to predict the residual error of an existing physics-based model, which finally results in a hybrid model. Those models have the advantage of being fully differentiable and natively implemented in an automatic differentiation framework, which means that their associated cost can be differentiated automatically via the chain rule. Overall, linking data assimilation and machine learning is a very hot topic, which has been recently reviewed in~\cite{ML-DA}.

\section{Proposed methods}
\label{Methods}

\subsection{Neural network design and training}
\label{Training}

In this section, we design a data-driven model that, when properly trained, should produce stable predictions on the long term.
A good indicator for the stability of long run predictions is that the eigenvalues of the learnt Koopman matrix should be located on the unit circle, which may encourage us to look for matrices with such eigenvalues. 
Among those are orthogonal matrices, which have many desirable properties. 
Most importantly, they constrain the dynamics to be periodic, as shown in the following result:\\

\begin{theorem}
[Discrete linear systems with special orthogonal matrices lead to periodic dynamics]
Let $\mathbf{K} \in \mathcal{SO}(d)$, the special orthogonal group of real invertible matrices satisfying $\mathbf{KK}^T = \mathbf{K}^T\mathbf{K} =  \mathbf{I}$ and with determinant equal to $+1$, and define a discrete-time dynamical system by 
\begin{equation}
    \mathbf{z}_{t+1} = \mathbf{Kz}_t
\end{equation}
with any initial condition $\mathbf{z}_0 \in \mathbb{R}^d$. Then there exists a continuous-time dynamical system 
\begin{equation}
    \frac{d\mathbf{z}}{dt} = \mathbf{Lz}
\end{equation}
with $\mathbf{z}(0) = \mathbf{z}_0$, and $\mathbf{L}$ a skew-symmetric matrix such that $\exp(\mathbf{L}) = \mathbf{K}$. Besides, the dynamics are periodic, i.e. $\exists \tau \in \mathbb{R}^+, \forall t \in \mathbb{R}^+, \mathbf{z}(t+\tau) = \mathbf{z}(t)$.\\
\label{thm:periodic}
\end{theorem}
The proof is relegated to Appendix~\ref{orthogonality}. This theorem shows that the dynamics of linear dynamical system specified with a skew-symmetric matrix (when continuous) or with a special orthogonal matrix (when discrete) leads to periodic dynamics. Note that this property carries on to any time independent transformation of $\mathbf{z}_t$: for any function $\psi$, $\psi(\mathbf{z}(t))$ will itself be periodic with the same period as $\mathbf{z}(t)$. 

\begin{figure}
    \centering
    \includegraphics[width=8.5cm]{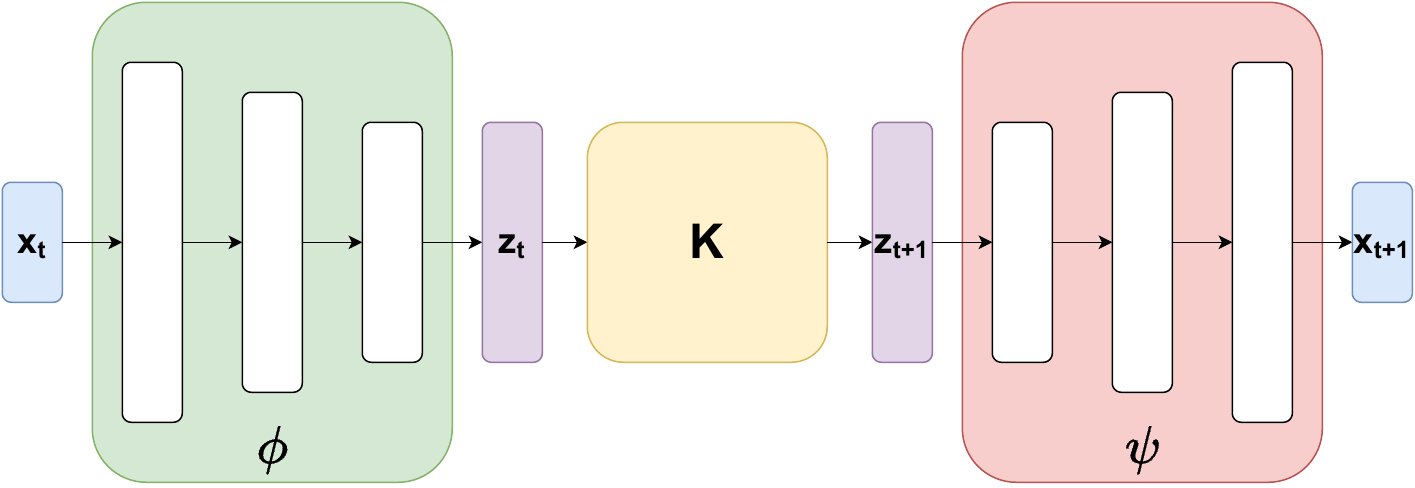}
    \caption{Schematic view of our architecture.}
    \label{fig:KoopmanAE}
\end{figure}

Suppose that we are modeling a dynamical system from which the state space is $\mathcal{D} \subset \mathbb{R}^n.$
Our architecture for a neural Koopman operator relies on three components: an encoding neural network $\phi$ with a decoder $\psi$ and a matrix $\mathbf{K} \in \mathbb{R}^{d \times d}$. It is graphically represented in Figure~\ref{fig:KoopmanAE}. The idea is that $(\phi, \psi)$ learns the relationship between the state space and a learnt $d$-dimensional (approximately) Koopman invariant subspace, while $\mathbf{K}$ corresponds to the restriction of the Koopman operator to this space.
Therefore, $\phi$ is expected to be a set of observation functions which constitute a Koopman invariant subspace. From now on, we will denote the latent variable corresponding to a vector $\mathbf{x}_t \in \mathcal{D}$ of the state space as $\mathbf{z}_t = \phi(\mathbf{x}_t)$.
Our goal is to be able to make long-term predictions of the state by successive multiplications of the encoded initial state, followed by a decoding step to come back to the original data space. This translates in equations as
\begin{equation}
\label{KoopmanPrediction}
    \psi(\mathbf{K}^{\tau}\phi(\mathbf{x}_t)) = \mathbf{x}_{t+ \tau}
\end{equation}
for any initial condition $\mathbf{x}_t$ and time increment $\tau$. We emphasize that $\tau$ does not necessarily have to be an integer since one can easily compute noninteger powers of $\mathbf{K}$ by using its matrix logarithm, as explained in section~\ref{Irregular}. The time increment $\tau$ could also very well be negative, enabling to predict the past state of a dynamical system from future states. A recent study~\cite{fathi2023course} suggests that it could be beneficial to periodically re-encode the latent vector $\mathbf{z}_{t+kp}$ with a chosen period $p$, but we do not explore this idea here.

Our training data will be constituted from $N$ time series of length $T$, which we denote as $(\mathbf{x}_{i,t})_{1\leq i \leq N, 0\leq t \leq T+1}$. Note that these time series could be possibly overlapping cuts of longer time series. A first processing step is to augment the state space with its discrete derivatives ${\mathbf{x}_{i,t} - \mathbf{x}_{i,t-1}}$. This means working with the variable $\mathbf{y}$ defined as 
\begin{equation}
\label{AugmentedState}
    \mathbf{y}_{i,t} = 
    \begin{pmatrix}
    \mathbf{x}_{i,t+1} & \mathbf{x}_{i,t+1} - \mathbf{x}_{i,t}
    \end{pmatrix}
    ^T
\end{equation}
for index $t \in [\![0,T]\!]$. This reformulation makes it easier to predict the future state. Indeed, given that the data varies smoothly, one could expect that $\mathbf{x}_{i,t} + (\mathbf{x}_{i,t} - \mathbf{x}_{i,t-1})$ is a good approximation of $\mathbf{x}_{i,t+1}$ (this formally looks like an explicit Euler scheme to integrate an underlying infinitesimal representation formulated as an ODE). This intuition is further theoretically justified by Takens' theorem~\cite{Takens}, which, informally, states that the evolution of a dynamical system gets more and more predictable when we know more time lags from an observed variable of the system. Using this augmented state is therefore useful when the observed $\mathbf{x}$ is not the state variable of the system. In practice, we work either with $\mathbf{x}$ or with $\mathbf{y}$ depending on what is possible and on the considered dynamical system. Since this choice does not influence our methods, we will always use the variable $\mathbf{x}$ in the following.

We denote by $\Theta$ the set of all the trainable parameters of our architecture. $\Theta$ includes the coefficients of $\mathbf{K}$ along with the trainable parameters of $\phi$ and $\psi$.
In order to obtain the desired behavior corresponding to equation~\eqref{KoopmanPrediction}, we train the architecture using the following loss terms:
\begin{itemize}
    \item The prediction term $L_{pred}$ ensures that the long-term predictions starting from the beginning of each time series are approximately correct. Some works~\cite{LRAN} weigh this loss with an exponentially decaying factor that gives more importance to short term predictions, but we choose to penalize the errors on all time spans equally:
    \begin{equation}
    \label{pred}
        \hspace{-1mm}
        L_{pred}(\Theta) = 
        \sum_{1\leq i \leq N}
        \sum_{1 \leq \tau \leq T}  
        ||\mathbf{x}_{i,\tau} - \psi(\mathbf{K}^{\tau}\phi(\mathbf{x}_{i,0}))||^2.
    \end{equation}
    \item The auto-encoding term $L_{ae}$ is the classical loss for auto-encoders, making sure that $\psi \circ \phi$ is close to the identity:
    \begin{equation}
    \label{ae}
    \hspace{-1mm}
        L_{ae}(\Theta) = 
        \sum_{1 \leq i \leq N}
        \sum_{0 \leq t \leq T}
        ||\mathbf{x}_{i,t} - \psi(\phi(\mathbf{x}_{i,t}))||^2.
    \end{equation}
    
    \item The linearity term $L_{lin}$ favors the linearity of the learnt latent dynamics. It is qualitatively similar to the residual in the formulation of EDMD in equation~\eqref{eq:EDMD}.
    \begin{equation}
    \label{lin}
        \hspace{-1mm}
        L_{lin}(\Theta) = 
        \sum_{1 \leq i \leq N}
        \sum_{1 \leq \tau \leq T}
        ||\phi(\mathbf{x}_{i,\tau}) - \mathbf{K}^{\tau}\phi(\mathbf{x}_{i,0})||^2.
    \end{equation}
    \item The orthogonality term is a regularisation term, prompting the complex eigenvalues of $\mathbf{K}$ to be located close to the unit circle, which favors the long-term stability of the latent predictions. It is particularly helpful when the dynamics are close to periodic, as shown by theorem~\ref{thm:periodic}. $||.||_F$ denotes the Frobenius norm.
    \begin{equation}
    \label{orth}
        L_{orth}(\mathbf{K}) = ||\mathbf{KK}^T - \mathbf{I}||_{F}^2.
    \end{equation}
\end{itemize}

As we will show throughout our experiments, the orthogonality term is very useful in self-supervised learning contexts, in the sense that it enables to train models that generalise better to downstream tasks such as interpolation, extrapolation and backward prediction. However, it has a more limited interest for supervised learning, and it can even be detrimental when used for modelling dynamical systems which have no seasonal component, which is why it should not always be included.

Note that, with our formulation, unlike with some more classical methods like DMD, it is hard to determine a minimal dimension $d$ required to model a dynamical system.
In practice, $d$ should be seen as a hyperparameter of the model, which will underfit the training data if it is too low and overfit the training data if it is too large. 

\subsection{Handling irregular time series}
\label{Irregular}

When working with irregular time series, it is not possible to augment the state with delayed observations as described in equation~\eqref{AugmentedState}, so that we necessarily use the input data $\mathbf{x}$ and not the augmented state $\mathbf{y}$. Yet, the training can still be performed in a way similar to the case of regular time series.
One has to distinguish two cases: (1) the data has a regular sampling with missing values (i.e. all temporal distances are multiples of a reference duration) and (2) the time increments between the sampled points are completely arbitrary.

If the irregular time series result from a regular sampling with missing values, then one can denote these data by $(\mathbf{x}_{i,t})_{1\leq i \leq N, 1\leq t \leq T}$, with the binary observation variable $(\mathbf{H}_{i,t})_{1\leq i \leq N, 1\leq t \leq T}$ being so that $\mathbf{H}_{i,t} = 1$ if $\mathbf{x}_{i,t}$ is actually observed and $0$ otherwise.
Then, one can trivially multiply each term of the prediction, auto-encoding and linearity losses from equations \eqref{pred}-\eqref{lin} by the corresponding binary coefficient $\mathbf{H}_{i,t}$ to train a model for these irregular data.

When the data is sampled at arbitrary times, one has to adopt a continuous formulation. In this case, one does not work with the discrete $\mathbf{K}$ but rather with its continuous counterpart $\mathbf{L}$, which is related to it through the matrix exponential 
\begin{equation}
\label{K-D}
    \mathbf{K} = \exp(\mathbf{L})
\end{equation} 
and can be seen as its corresponding infinitesimal evolution. A sufficient condition to guarantee the existence of such a matrix $\mathbf{L}$ is that $\mathbf{K}$ (almost always diagonalizable in $\mathbb{C}$) has no real negative eigenvalue~\cite{MatrixLog}. In our case, we constrain $\mathbf{K}$ to be close to orthogonal and initialize it to the identity. Thus, the eigenvalues are very unlikely to become real negative. Indeed, the set of negative real numbers has zero Lebesgue measure. This means that the next noisy iterate provided by a stochastic gradient descent (or related algorithms) will perturb the eigenvalues, and if the associated eigenvalue distribution admits a density w.r.t. the Lebesgue measure, the eigenvalues will almost surely avoid this set, unless it is an attractor. At any rate, this never happened in our experiments.

Under this assumption, we can equivalently switch to a continuous dynamical system whose evolution can be described in a Koopman invariant subspace by
\begin{equation}
\label{infinitesimal}
    \frac{d\phi(\mathbf{x}(t))}{dt} = \frac{d\mathbf{z}(t)}{dt} = \mathbf{L}\mathbf{z}(t).
\end{equation}
In this case, it is a well known result that
\begin{equation}
\label{Continuous}
    \mathbf{z}(t_0+\tau) = \exp(\tau \mathbf{L}) \mathbf{z}(t_0) 
\end{equation}
for any time increment $\tau \in \mathbb{R}$. In particular, with $\tau=1$, we find the previous definition of $\mathbf{K}$ from equation~\eqref{K-D}.

Let us suppose that we train a model on $N$ irregular time series. For each index $1 \leq i \leq N$, we denote the trajectory $\mathbf{x}_i$ as a list of $T_i$ time-value pairs $(t_{i, k},\mathbf{x}_{i,k})_{0 \leq k \leq T_i}$. Without loss of generality, one can suppose that the pairs are ordered by increasing times, with $t_{i,0} = 0$. The set of trainable parameters $\Theta$ now includes the parameters of $(\phi,\psi)$ and the coefficients of the infinitesimal evolution matrix $\mathbf{L}$. Then, one can rewrite the prediction, auto-encoding and linearity loss terms as:
\begin{equation}
    L_{pred}(\Theta) = \sum_{1 \leq i \leq N} \sum_{1\leq k \leq T_i} ||\mathbf{x}_{i,k} - \psi(\mathbf{K}^{t_{i,k}}\phi(\mathbf{x}_{i,0}))||^2
\end{equation}
\begin{equation}
    L_{ae}(\Theta) = \sum_{1 \leq i \leq N} \sum_{0\leq k \leq T_i} ||\mathbf{x}_{i,k} - \psi(\phi(\mathbf{x}_{i,k}))||^2
\end{equation}
\begin{equation}
    L_{lin}(\Theta) = \sum_{1 \leq i \leq N} \sum_{1\leq k \leq T_i} ||\phi(\mathbf{x}_{i,k}) - \mathbf{K}^{t_{i,k}}\phi(\mathbf{x}_{i,0})||^2
\end{equation}
where we use the slightly abusive notation $\mathbf{K}^t = \exp(t\mathbf{L})$ for any non-integer time increment $t$. Now, one can use these rewritten loss terms in conjunction with the unchanged orthogonality loss to learn from irregularly-sampled data in the same way as from regularly-sampled ones, although it is likely to be a more challenging learning problem.

The continuous formulation is the most general one and it can also be used when the training data are regularly sampled. However, this requires computing the matrix exponential $\mathbf{K} = \exp(\mathbf{L})$ after each update of $\mathbf{L}$ instead of working on $\mathbf{K}$ directly. 
We experimentally found that, when training on regularly-sampled data, the discrete formulation has slightly better performances than the continuous one.
We conjecture that this is due to the gradient of the loss function being more complex when performing a matrix exponential, and we recommend using the continuous formulation only when necessary.
Note however that training a model with a discrete formulation does not mean giving up on the continuous modelling. Indeed, when one has a trained discrete matrix of evolution $\mathbf{K}$ at hand, it is possible to switch to continuous dynamics as soon as a matrix logarithm exists~\cite{MatrixLog}. In that case, the complex eigendecomposition of $\mathbf{K}$ writes
\begin{equation}
    \mathbf{K} = \mathbf{V} \mathbf{\Lambda} \mathbf{V}^{-1}
\end{equation}
with $\mathbf{V} \in \mathbb{C}^{d\times d}$ and $\mathbf{\Lambda} \in \mathbb{C}^{d \times d}$ a diagonal matrix. Then, $\mathbf{L}$ can be obtained by computing the principal logarithm of each (necessarily not real negative) diagonal coefficient of $\mathbf{\Lambda}$:
\begin{equation}
    \mathbf{L} = \mathbf{V}\log(\mathbf{\Lambda})\mathbf{V}^{-1}.
\end{equation}
One can easily check that $\mathbf{L}$ then verifies equation~\eqref{K-D}, and use this matrix to query the state of the latent system at any time from a given initial condition using equation~\eqref{Continuous}.

\subsection{Variational data assimilation using our trained model}
\label{Assimilation}

Once a model has been trained for a simple prediction task, it is supposed to hold enough information to help solve a variety of inverse problems involving the dynamics, like interpolation or denoising. To leverage this knowledge, we resort to variational data assimilation, using the trained model as a dynamical prior instead of a more classical hand-crafted physical prior. 
We describe hereafter a general formulation for inverse problems involving time series of images  and different methods to solve them. 
Although we consider images specifically in our experiments, the methods can be used for any time series by ignoring or adapting the spatial prior.

Suppose that we are working on images containing $N$ pixels and $L$ spectral bands ($L$ being 3 for RGB images or higher for multi/hyperspectral images), defined on a set of $T$ time steps with some missing values.
We denote this data by $(\mathbf{\Tilde{x}}_t)_{t\in H}$ with $H\subset [\![0,T ]\!]$. 
For each $t \in H$, $\mathbf{\Tilde{x}}_t \in \mathbb{R}^{N \times L}$. As is classical in data assimilation, we assume that $\mathbf{\Tilde{x}}$ is a noisy and incomplete version of an unknown underlying truth $\mathbf{x}^{true} = (\mathbf{x}_0^{true}, ..., \mathbf{x}_T^{true})$.
Our objective is to reconstruct (and possibly extend) a complete time series $\mathbf{x} \in \mathbb{R}^{(T+1) \times N \times L}$ that is expected to approximate $\mathbf{x}^{true}$.

The first method that we propose is a weakly-constrained variational data assimilation, where we minimise a variational cost on $\mathbf{x}$ which is composed of at most three components: a term of fidelity to the available data, a dynamical prior which is given by our model, and a spatial prior. The variational cost on $\mathbf{x}$ can thus be expressed as
\begin{equation}
\label{classicalDAequation}
    \sum_{t \in H} ||\Tilde{\mathbf{x}}_t - \mathbf{x}_t||^2 + \alpha\sum_{t=0}^{T-1} ||\mathbf{x}_{t+1} - \mathcal{M}(\mathbf{x}_{t})||^2 + \beta S(\mathbf{x})
\end{equation}
where $\mathcal{M}(\mathbf{x}_{t}) = \psi(\mathbf{K}\phi(\mathbf{x}_{t}))$ and $S$ is the spatial prior. In practice, $S$ can be a classical spatial regularisation leading to spatially smooth images, such as a Tikhonov regularisation~\cite{Tikhonov} or the total variation~\cite{TV}. We emphasize that the optimised variable here is the whole time series $\mathbf{x}$.
The first term of equation~\eqref{classicalDAequation} is the data fidelity term (first term of equation~\eqref{GeneralDAEquation}) and the other two terms form together the prior or regularisation term (second term of equation~\eqref{GeneralDAEquation}).

In some cases, it can be useful to consider a more constrained optimisation. This is especially true when dealing with very noisy data, in which case the data fidelity term can lead to overfitting the noise even if a high weight is put on the prior terms. 
In such cases, we do not optimise on $\mathbf{x}$ anymore but rather on the latent initial state $\mathbf{z}_0$ of the prediction, so that only values of $\mathbf{x}$ that can be produced by our data-driven dynamical prior are considered.
In this way, we seek to solve
\begin{equation}
\label{ConstrainedDAequation}
    \mathbf{z}_0^* = 
    \underset{\mathbf{z}_0 \in \mathbb{R}^{N \times d}}{\textrm{arg min}}\sum_{t \in H} ||\Tilde{\mathbf{x}}_{t} - \mathbf{x}_t(\mathbf{z}_0)||^2 + \beta S(\mathbf{x}(\mathbf{z}_0)),
\end{equation}
where, for time $t$, $\mathbf{x}_t(\mathbf{z}_0) = \psi(\mathbf{K}^{t} \mathbf{z}_0)$. After finding the optimal initial condition $\mathbf{z}^*_0$, one can simply compute the associated predictions at any time $t$ using $\mathbf{x}_t(\mathbf{z}_0^*)$. Note that $\mathbf{z}_0$ belongs to $\mathbb{R}^{N \times d}$ since we assumed that the input of $\phi$ is the reflectance vector of a single pixel of an image, so that the model forecasts the dynamics of all pixels in parallel.

Although it might seem from equation~\eqref{ConstrainedDAequation} that the encoder $\phi$ of our model is of no use here, it is actually important in two ways. First, it is necessary for training the other components $\psi$ and $\mathbf{K}$, and in particular for ensuring that $\mathbf{x}_t(\mathbf{z}_0)$ is always a realistic state of the system. Second, the optimisation problem is initialised with $\mathbf{z}_0 = \phi(\mathbf{x}_0)$, which is a logical guess that helps boosting the performance since this problem is not convex and is solved by automatic differentiation.

The constrained and unconstrained assimilations have different advantages and weaknesses. The unconstrained assimilation is generally useful when $\mathbf{\Tilde{x}}$ is close to $\mathbf{x}^{true}$, i.e. when the observed data are complete and when the signal-to-noise ratio is large enough. In this case, it can efficiently provide small corrections to the noise in the dynamics. The constrained optimisation, however, is not useful in this case since it is not able to reconstruct the exact observed data. Therefore, when the error due to the noise has a magnitude smaller or similar to the reconstruction error of the model, the constrained optimisation is not able to perform a relevant denoising. On the other hand, the constrained assimilation is very effective to deal with sparse or very noisy observed data. Indeed, any prediction made by the model should be a possible trajectory of the dynamical system, which means that the result of this optimisation can be seen in some way as the plausible trajectory which best matches the observed data.

Another possibility in our framework is to perform a constrained assimilation like in equation~\eqref{ConstrainedDAequation} but with a joint optimisation of the initial latent space and of the parameters of the model. Thus, the optimisation problem becomes
\begin{equation}
\label{ConstrainedDAEquation2}
    \underset{\mathbf{z}_0, \mathbf{K}, \psi}{\textrm{min}}\sum_{t \in H} ||\Tilde{\mathbf{x}}_{t} - \mathbf{x}_t(\mathbf{z}_0)||^2 + \beta S(\mathbf{x}(\mathbf{z}_0)),
\end{equation}
where, again, $\mathbf{x}_t(\mathbf{z}_0) = \psi(\mathbf{K}^{t} \mathbf{z}_0)$. While this problem would be very difficult to solve when starting with random variable initialisations, using the parameters of a pretrained model as initial values gives good results. Notably, adjusting the model parameters enables to fit the available data much better, which is especially useful when working on a set of data which differs from the model training data. This strategy corresponds, in machine learning parlance, to transfer learning or solving a downstream task using the self-supervised trained model. However, one must be careful not to make the model overfit the assimilated data at the cost of its general knowledge, which could be related to the well-known catastrophic forgetting~\cite{CatastrophicForgetting}. 
Thus, it is critical to use a very low learning rate as commonly described in the literature, e.g.~\cite{FineTuningBert}.

\section{Experiments on simulated data}

\label{ExperimentsSimulated}

Here, we present a benchmark of our method against the methods of~\cite{Lusch} and~\cite{Azencot}, which we respectively call DeepKoopman and consistent Koopman autoencoder (cKAE) on a 3-dimensional dynamical system arising from fluid dynamics.

The nonlinear fluid flow past a cylinder with a Reynolds number of 100 has been a fluid dynamics benchmark for decades, and it was proven by~\cite{Noack} that its high-dimensional dynamics evolves on
a 3-dimensional attractor with the model:
\begin{equation}
\label{FluidFlowEquations}
\begin{aligned}
    &\dot{x} = \mu x - y - xz\\
    &\dot{y} = \mu y + x - yz\\
    &\dot{z} = - y + x^2 + y^2.
\end{aligned}
\end{equation}
This dynamical system is not periodic, yet it exhibits a stable limit cycle and an unstable equilibrium.

In our experiments, we use the training and test data from~\cite{Lusch}, which have been generated by numerically integrating equations~\eqref{FluidFlowEquations}. All of our models trained on this dynamical system have the same architecture: the encoder $\phi$ (resp. decoder $\psi$) is a Multi-Layer Perceptron (MLP) with 2 hidden layers of size 256 and 128 (resp. 128 and 256), with the dimension of the latent space and matrix $\mathbf{K}$ being $d=16$. We use the same hyperparameters for cKAE, which also includes a backward evolution matrix $\mathbf{D}$ of the same size as $\mathbf{K}$. For DeepKoopman, we use the hyperparameters reported in~\cite{Lusch}.
\subsection{Interpolation from low-frequency regular data}
\label{Low-freq}

We first show the ability of our architecture to model a continuous dynamical system even when trained on discrete data. As mentioned in subsection~\ref{Irregular}, once a model is trained, one can retrieve its learnt Koopman matrix $\mathbf{K}$ and compute its corresponding infinitesimal operator $\mathbf{L}$. Then, one can analytically compute a new discrete matrix corresponding to an advancement by any desired time increment. Concretely, suppose that $\bar{\omega}$ is a chosen target frequency (we choose $\bar{\omega} = 50$~Hz here).
We train a model on training time series sampled at a lower frequency $\omega$, obtaining in particular a Koopman matrix $\mathbf{K}$. Then, we compute the discrete operator $\bar{\mathbf{K}} = \exp{(\frac{\omega}{\Bar{\omega}}\log{\mathbf{K}})} = \exp{(\frac{\omega}{\Bar{\omega}}\mathbf{L})}$, and use it to perform predictions at a frequency $\Bar{\omega}$ from the initial states of the test time series. Finally, we compute a mean squared error (MSE) between the high-frequency groundtruth and predictions, averaged over all time steps and trajectories from the test set.

Note that, for this experiment, our models are trained without the orthogonality loss term from equation~\eqref{orth} since the considered dynamical system at hand, having no quasi-periodic component, is not well adapted for this regularisation. We compare our results against DeepKoopman and cKAE models trained on the same data and for which we similarly computed $\bar{\mathbf{K}}$. This was not proposed in their original papers, but a more naive interpolation method, such as a linear interpolation in the latent space, yields unsatisfactory results. Note that both cKAE and our model feature a single matrix $\mathbf{K}$, and therefore only one matrix $\bar{\mathbf{K}}$ needs to be computed for these models. In contrast, DeepKoopman computes a new matrix $\mathbf{K}$ for each state and time increment, and one must compute a new $\Bar{\mathbf{K}}$ for each of those, leading to a far greater amount of computation at inference. In addition, training the DeepKoopman model is significantly more costly than the other models due to the necessity to compute a new matrix $\mathbf{K}$ through the auxiliary network at each time step. cKAE is itself slower to train than our model because of its consistency loss from equation~\eqref{eq:consistencyLoss}, which requires computing $2d$ additional matrix multiplications for each training step. Concretely, in our configuration, the duration of a training epoch in frequency $\omega = 50$~Hz is approximately 3.4 seconds for DeepKoopman, 1 second for cKAE and 0.9 second for our model.

We report in Table~\ref{VaryingFrequencyTable} the results obtained with various training frequencies $\omega$. All trainings are repeated 5 times with different initialisations of the model's parameters, and we report the mean and standard deviation of the errors over these 5 runs. The results show that the quality of the high-frequency predictions of our model depends very little on the training frequency.
The MSE for DeepKoopman and cKAE are similar to ours for the highest training frequencies but increase faster as the training frequency decreases. We emphasize that all instances of all models excel at predicting in their training frequency, with DeepKoopman being the best model for any frequency. The main difficulty in this experiment is to obtain models that are also good at interpolating to a high frequency. In this regard, at training frequencies $5$~Hz and $2.5$~Hz, some instances of DeepKoopman achieve a relative success with a MSE in the order of $10^{-6}$ while some remarkably fail with a MSE in the order of $10^{-3}$ or $10^{-4}$ (hence the very high standard deviations). However, none of the instances is as good as the mean performance of our model for these frequencies. cKAE, in contrast, consistently fails to interpolate in the lowest frequencies. The main reason for this failure seems to be the absence of a linearity term in their loss function. Indeed, a quick (unreported) experiment enabled us to obtain an error in the order of $10^{-6}$ with training frequency $2.5$~Hz by training a cKAE instance with an additional forward linearity loss term.

\begin{figure}
    \centering
    \includegraphics[width=8.5cm]{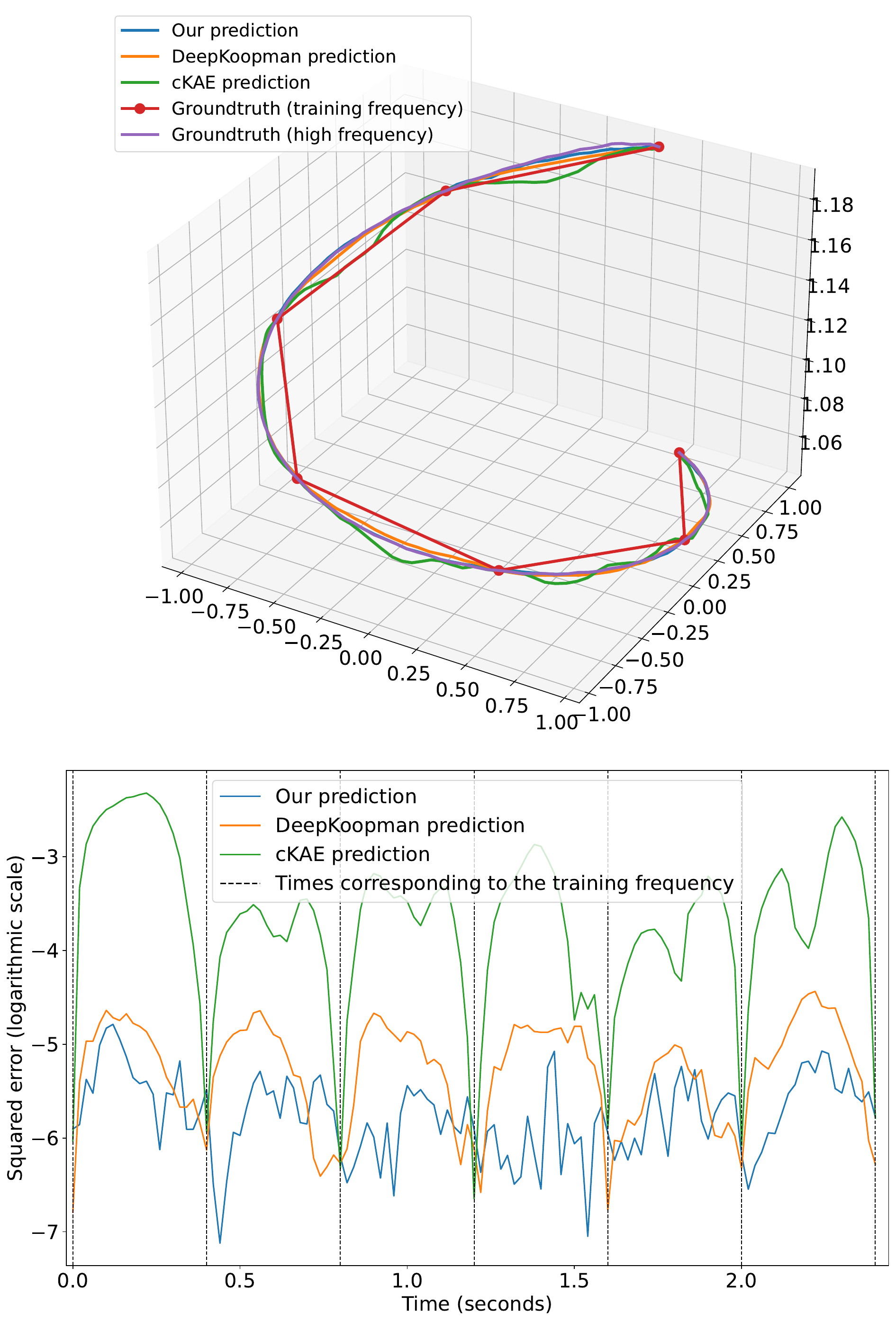}
    \caption{Upsampling experiment on fluid flow data. We learn a model on low-frequency data and then use the continuous representation to make a high-frequency prediction which we compare to the groundtruth on test trajectories. Top: the three tested models compared to a groundtruth trajectory. Bottom: corresponding mean squared errors over time, in a logarithmic scale.}
    \label{fig:Upsampling}
\end{figure}

Our model, however, successfully combines the information of many low-resolution time series to construct a faithful continuous representation of the dynamics. On Figure~\ref{fig:Upsampling}, we show the interpolation by the best instance trained at $\omega = 2.5$~Hz for each of the three models on a test trajectory. It appears that, although all models are able to fit the groundtruth points corresponding to the low training frequency, they are not all able to interpolate to a higher frequency. Only our model performs as well for interpolating than for fitting the points corresponding to its training frequency.

\begin{table}[htbp]
    \centering

\caption[size=9pt]{Mean squared errors ($\times 10^{-6}$) for 50~Hz interpolation of fluid flow dynamics with various training frequencies}
\label{VaryingFrequencyTable}
\begin{tabular}{|c|c|c|c|}

\hline
Training & DeepKoopman & cKAE & Our method \\
frequency &  &  &  \\
\hline
50 Hz & $\mathbf{1.16 \pm 0.05}$ & $1.31 \pm 0.23$ & $1.51 \pm 0.19$ \\
\hline
25 Hz & $1.30 \pm 0.09$ & $1.84 \pm 0.57$ & $\mathbf{1.20 \pm 0.13}$ \\
\hline
10 Hz & $1.59 \pm 0.01$ & $2.92 \pm 1.01 $ & $\mathbf{1.42 \pm 0.19}$ \\
\hline
5 Hz & $37.4 \pm 44.2$ & $18.7 \pm 6.25$ & $\mathbf{1.56 \pm 0.03}$ \\
\hline
2.5 Hz & $294 \pm 578$ & $971 \pm 403$ & $\mathbf{2.02 \pm 0.42}$ \\
\hline
\end{tabular}
\end{table}
\label{SamplingRateTable}

\subsection{Learning to predict forward and backward} 
\label{Forward-backward}
\begin{figure}
    \centering
    \includegraphics[width=8.5cm]{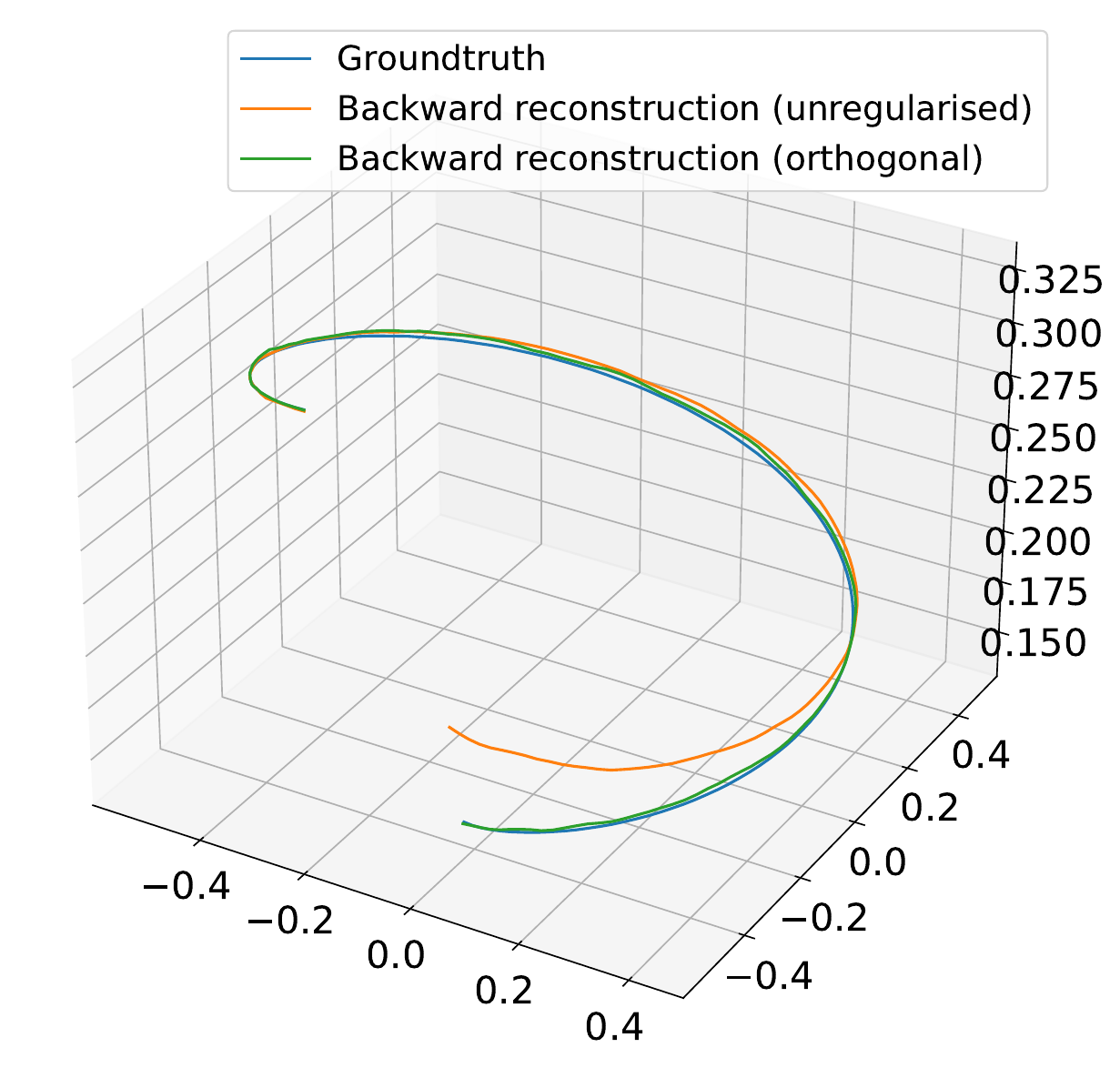}
    \caption{Backward reconstructions of a test time series from a model trained with an orthogonality loss term (orthogonal) and a model trained without it (unregularised). Both models were trained on high-frequency forecasting.}
    \label{fig:Backward}
\end{figure}

Here, we investigate the ability of our model to perform backward predictions after being trained on forward prediction. In theory, one can simply invert the learnt matrix $\mathbf{K}$. This is not a possibility of the DeepKoopman framework since it computes a new matrix $\mathbf{K}$ as a function of the input at each iteration, and one would need to invert the matrix of the preceding state (which one does not have access to) to predict backwards. We test the performance of our models in this context and study the influence of using the regularising orthogonality loss term from equation~\eqref{orth} against not using it (we remind that we did not use it in the interpolation experiment above).Table~\ref{ForwardBackwardTable} reports our mean squared errors: in this table, "HF" means that the model was trained on high-frequency data (50~Hz) while "LF" means that it was trained on low-frequency data (2.5~Hz).  
One can see that the backward predictions have significantly higher errors in average than forward predictions. This matches the observation by~\cite{Azencot} that naively inverting a learnt Koopman matrix is generally not effective. In particular, our model trained on low-frequency forecasting with no orthogonality term quickly diverges from the groundtruth when performing backward predictions. However, our model with an orthogonality regularisation trades reduced forecasting performance for the ability to run better backward reconstructions although it was not trained on this task. 
Figure~\ref{fig:Backward} shows a typical example for which the model with an orthogonal matrix sticks to the time series while the unregularised one diverges from it.
Interestingly, we noticed than using the inverse of the matrix from our model trained with an orthogonality loss worked better than using the inverse of the forward prediction matrix of a trained cKAE model. This is somewhat surprising since the cKAE model features a consistency loss that should make its forward prediction model better at backward predictions than a model that is completely agnostic to the backward dynamics. However, the orthogonality loss seems to perform an unexpectedly strong regularisation in this context.


\begin{table}[htbp]
    \centering

\caption[size=9pt]{Mean squared errors for forward and backward predictions}
\label{ForwardBackwardTable}
\begin{tabular}{|c|c|c|}

\hline
Task \textbackslash \hspace{0.1cm} Model & Unregularised & With orthogonality \\
\hline
Forward prediction (HF) & $\mathbf{1.65\times10^{-6}}$ & $5.50 \times 10^{-6}$ \\
\hline
Backward prediction (HF) & $1.88 \times 10^{-4}$ & $\mathbf{1.48 \times 10^{-4}}$ \\
\hline
Forward prediction (LF) & $\mathbf{1.66 \times 10^{-6}}$ & $8.89 \times 10^{-6}$ \\
\hline
Backward prediction (LF) & $0.123$ & $\mathbf{3.24 \times 10^{-5}}$ \\
\hline
\end{tabular}
\end{table}

\section{Experiments on real Sentinel-2 time series}
\label{ExperimentsReal}

\begin{figure}
    \centering
    \includegraphics[width=8.5cm]{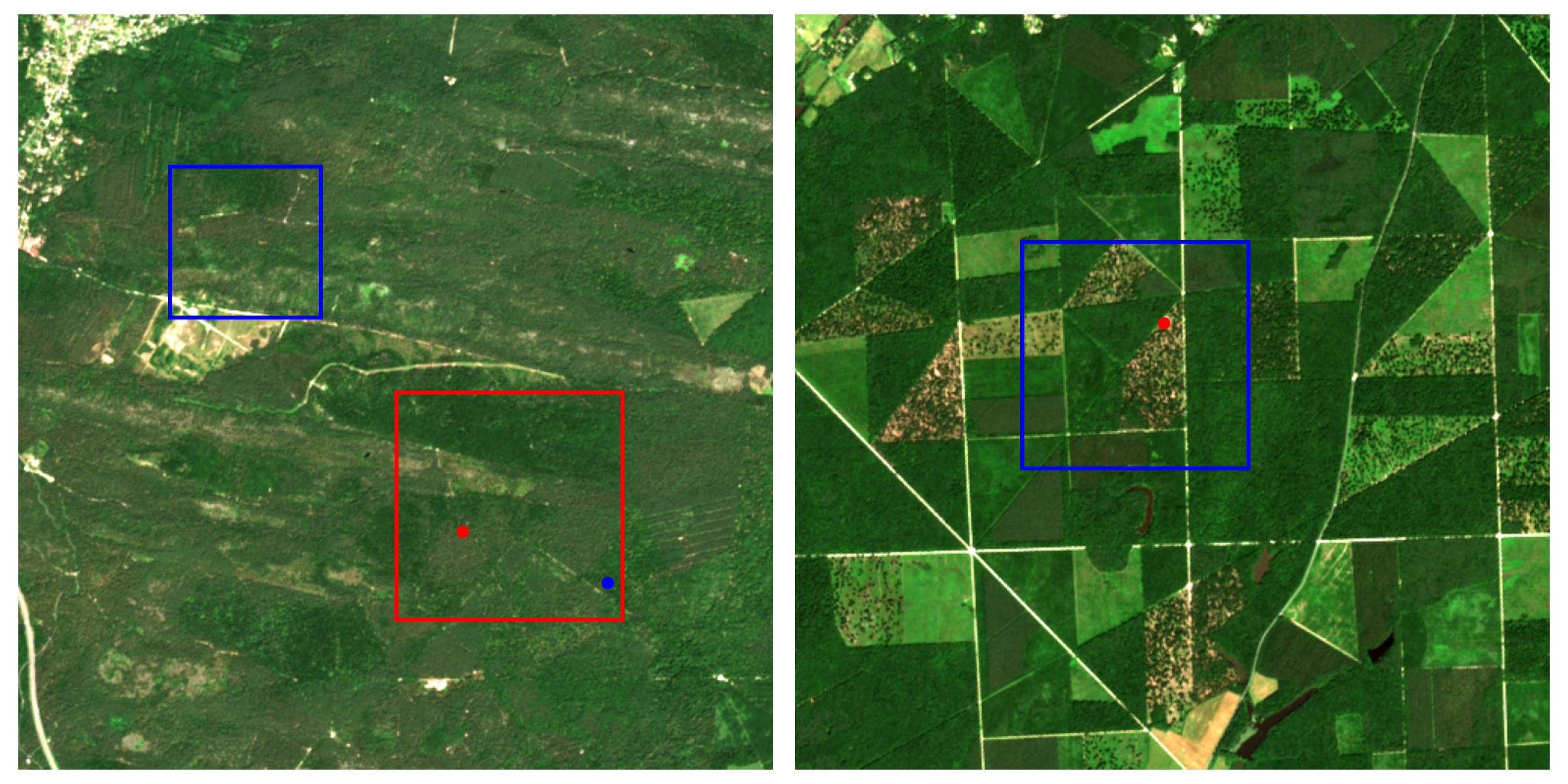}
    \caption{Left: a temporally interpolated Fontainebleau image. Right: a non-
interpolated Orléans image. The date for both images is 20/06/2018. Those
are RGB compositions with saturated colors. The red square is the $150 \times 150$ pixel training area and the blue squares are test areas.}
    \label{fig:data_samples_Sentinel}
\end{figure}

In this section, we work on multispectral satellite image time series. They consist in successive multivariate images of the forests of Fontainebleau and Orléans in France, which have been taken by the Sentinel-2 satellites as a part of the European Copernicus program~\cite{Copernicus} over a duration of nearly 5 years. We use the reflectance from $L=10$ visible and infrared spectral bands at a spatial resolution of 10 meters, resorting to bicubic interpolation for those that were originally at a 20 meter resolution. Although the satellites have a revisit time of five days, many images are unexploitable due to the presence of too many clouds between the satellite and the surface. Therefore, we performed temporal interpolation of the available data to obtain complete versions of the time series.
The interpolation is performed with the Cressman method, which fills each missing value with a normalized sum of the available data, weighted by a Gaussian function of the temporal distance to the filled time, with a Gaussian radius of 15 days.
After this, we have 2 versions of each image time series at our disposal: one incomplete version where all data are real, and a complete but partly synthetic version. Having synthetic parts in a real-world time series is generally not desirable since a model trained on this data will learn the chosen interpolation scheme along with the true data distribution, but it can be necessary since most models are not able to handle irregularly-sampled time series. In practice, we will show that models trained on interpolated data can transfer well to raw incomplete data.
We show RGB compositions of sample images from the time series in figure~\ref{fig:data_samples_Sentinel}. Further details are available in~\cite{EUSIPCO}. The datasets, along with the code for the experiments, are freely accessible from \url{github.com/anthony-frion/Sentinel2TS}. Throughout this section, the reported mean squared errors are averaged over all spectral bands, pixels and available times.

\subsection{Forecasting}
\label{forecasting}

First, we train models to perform predictions from an initial condition on the $2L$-dimensional reflectance vector of a given pixel. The input dimension is twice the number of spectral bands of the pixels since we work with delayed embeddings as defined in equation~\eqref{AugmentedState}. Our encoder $\phi$ is a multi-layer perceptron with hidden layers of size 512 and 256, and the decoder $\psi$ symmetrically has hidden layers of size 256 and 512. The latent space and matrix $\mathbf{K}$ have a size of $d=32$.
We train models with this architecture with and without the orthogonality loss term from equation~\eqref{orth} in order to assess its influence on the trained models. These models are compared with a long short-term memory~\cite{LSTM} (LSTM) model with hidden size 256, which, like ours, has around $3 \times 10^{5}$ parameters. 
In addition, in order to assess the importance of our model's nonlinear embedding, we compare against a baseline linear model which assumes that $\phi$ and $\psi$ are simply identity functions in equation~\eqref{KoopmanPrediction}, thus consisting of only a matrix $\mathbf{K} \in \mathbb{R}^{2L \times 2L}$
This baseline is similar to dynamic mode decomposition~\cite{DMD}, yet we observed that computing $\mathbf{K}$ as the closed-form least-squares solution for 1-step prediction leads to poor long-term predictions, and we found that an accurate long-term reconstruction was only possible with a high order delayed embedding~\cite{HODMD}. To circumvent this issue, we trained this model to long-term prediction with automatic differentiation, like the other models of this benchmark. This model will be referred to hereafter as long term DMD.

The models are trained on the interpolated version of the Fontainebleau dataset, containing $T_{test}=342$ images, of which we use the first $T_{train}=242$ for training. The spatial area that is used is contained in the red square from figure~\ref{fig:data_samples_Sentinel}. Note that we do not train the models on $T_{train}$-time-steps pixel reflectance time series but on time series of length $T=100$ extracted from the data. In this way, the models learn to predict from any initial condition rather than just from the initial time of the dataset. 
The set of 100-time-steps time series that are used for training is fixed and organised in 512 batches of size 512, which are always passed through the models in a deterministic order. This enables to retrain the models and to obtain the exact same results as ours with the available code.
We use as the validation loss the mean-squared error of a $T_{train}$-step prediction from the state at time 1. Since the models tend to overfit to the $T$-{step} prediction task, we use the validation loss as an early stopping criterion, and always save the model that minimizes it. All of the studied models are trained with this procedure on the exact same data.

All models are evaluated on the task of predicting the state at times $T_{train}$ to $T_{test}$ using only one delayed embedding at time 1. Their performance are measured through the mean squared error (MSE) over all times, pixels and spectral bands in two areas: the Fontainebleau training area and a test area in the forest of Orléans, marked by a blue square in figure~\ref{fig:data_samples_Sentinel}. Thus, the Fontainebleau forecasting performance only reflects the ability of the models to extrapolate in time while the Orléans testing area also tests the ability of the model to transfer to different initial conditions. For the Fontainebleau area, we use the original irregular time series, which means that the models are tested only on true data. 
Since it is not possible to use a delayed embedding in this context, we make a rough estimate of the finite difference derivative from the first two available snapshots.
All results are displayed in table~\ref{ForecastingTable}. For all models but the long term DMD, we train 5 different instances with different random parameterisations, and we report the mean and standard deviation of the MSE over these 5 instances. We also show the MSE of the best instance from each model in parenthesis for both datasets. For long term DMD, we noticed that initialising $\mathbf{K}$ with the identity matrix gave better performance than any random initialisation, which is why we only report this result, with no standard deviation.

From the reported results, one can see that
the impact of the orthogonality loss is higher on the test Orléans area, which shows that this loss term is especially useful for transferring to new data. Our model with the orthogonality loss is the best on the Fontainebleau area and second best in the Orléans area. The LSTM model performs almost on par with ours on the training area, yet its performance are much worse when testing it on the test Orléans area. This can be explained by the fact that the underlying evolution of the LSTM is nonlinear, which makes it a more complex model which is more prone to overfitting on its training data. Long term DMD is able to roughly approximate the periodic pattern of the data, yet its limited capacity makes it unable to capture the precise relationship between the initial condition and the long-term behavior or to fit a complex periodic pattern. However, its simplicity also makes it a very good choice when transferring to a test area since it is not sensible to overfitting (relatively high bias, but also relatively low variance). It even outperforms the mean error of our model with orthogonality loss in this case: in detail, only 2 of the 5 instances of our model trained with an orthogonality regularisation outperform long term DMD on the test Orléans area in this setting. Indeed, long term DMD suffers less from the shift of the distribution of the data than the other models. However, we shall see next that our model is still a much better prior for data assimilation.

\begin{table}[htbp]
\centering
\caption[size=9pt]{Forecasting MSE ($\times 10^{-3}$) for different areas and methods}
\label{ForecastingTable}
\begin{tabular}{|c|c|c|}
\hline
 & Fontainebleau  & Orléans (irregular data) \\
\hline
Our model & $1.98 \pm 0.20$ & $10.35 \pm 0.97$ \\
(with orthogonality) & $(\mathbf{1.76})$ & $(\mathbf{8.93})$ \\
\hline
Our model & $2.15 \pm 0.28$ & $12.06 \pm 0.76$ \\
(no orthogonality) & $(1.85)$ & $(10.66)$ \\
\hline
LSTM & $2.08 \pm 0.11 $ & $12.78 \pm 1.62$ \\
 & $(1.99)$ & $(10.95)$ \\
\hline
Long term DMD & $4.32$ & $9.79$ \\
\hline
\end{tabular}
\end{table}

\subsection{Forecasting with data assimilation}
\label{AssimilationForecasting}

\begin{figure}
    \centering
    \includegraphics[width=8.5cm]{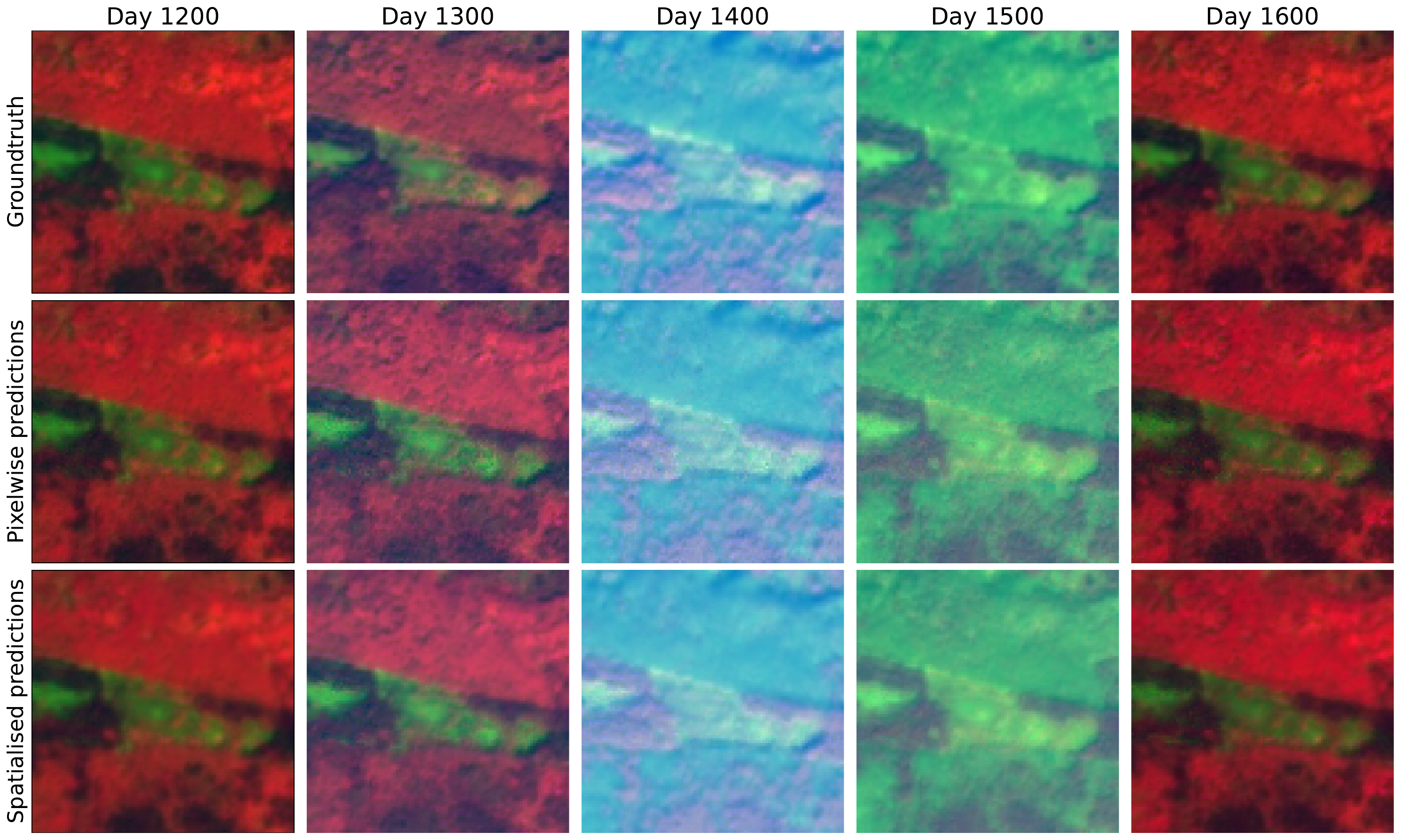}
    \caption{Top: groundtruth images of Fontainebleau, corresponding to test times. Middle: predictions made by our model by assimilating the time series up to day 1200 with a trained model. Bottom: Same as middle but including a spatial regularisation in the variational cost.
    The colors result from a 3-dimensional principal component analysis (PCA) of the 10 spectral bands performed 
    globally on all the Fontainebleau data.
    This is much more informative than an RGB composition, mainly because vegetation is very reflective in the near-infrared domain.}
    \label{fig:pred_Fontainebleau}
\end{figure}

We showed in the last section that a trained model is able to model the long-term reflectance dynamics of a pixel from only 2 observations (because of the delayed embedding). However, one can obtain a much more accurate forecast by taking into account a higher number of observations for a given pixel. Namely, one can try to predict the future dynamics of a pixel after time $T_{train}$ given a time series $(\Tilde{\mathbf{x}}_t)_{1 \leq t \leq T_{train}}$ of observed data representing its past behavior. We perform this task using the variational cost from equation~\eqref{ConstrainedDAequation}, where the set $H$ of observed time indices contains all positive integers up to time $T_{train} = 242$ (i.e. the training data).
We minimise this cost with no spatial regularisation or with a simple Tikhonov regularisation favoring the spatial smoothness of the resulting time series. More precisely, we seek to solve
\begin{equation}
\label{eq:AssimilationForecasting}
    \mathbf{z}_0^* = \underset{\mathbf{z}_0 \in \mathbb{R}^{N \times d}}{\textrm{arg min}}
    \sum_{t=0}^{T_{train}} 
    ||\mathbf{\Tilde{x}}_{t} - 
    \mathbf{x}_t(\mathbf{z}_0)||^2 + 
    \beta S(\mathbf{x}_t(\mathbf{z}_0))
\end{equation}
where we again note $\mathbf{x}_t(\mathbf{z}_0) = \psi(\mathbf{K}^{t} \mathbf{z}_0)$, and $S(\mathbf{x})$ is a smoothness prior, penalising the square of the spatial gradient of the resulting images through first order finite differences. 
The case with no spatial regularisation simply corresponds to $\beta = 0$. In this case, the optimisation variable $\mathbf{z_0} \in \mathbb{R}^{N \times d}$ can be seen as $N$ independent vectors which do not have to be computed in parallel, alleviating the memory requirements.
Once $\mathbf{z}_0^*$ has been computed, one can use it to extend the generated time series by simply using higher powers of $\mathbf{K}$. 
We will refer to this technique as assimilation-forecasting.
We solve this equation with gradient descent, using $\mathbf{z}_0 = \phi(\Tilde{\mathbf{x}}_0)$ as a starting point, which is more effective than a random starting point. Should $\mathbf{z}_0^*$ be equal to $\phi(\Tilde{\mathbf{x}}_0)$, then this would be equivalent to simply performing a forecast from the initial state $\Tilde{\mathbf{x}}_0$, which is not the case in practice.
One can observe assimilation-forecasting results of our model with and without spatial prior in a 3-dimensional PCA projection of a $100\times100$ subcrop of the data in figure~\ref{fig:pred_Fontainebleau}. Note that, even though all the pixels correspond to a forest environment, one can visually see that there are various long-term patterns among the pixels, and that our model can reproduce all of them. In addition, although the spatial prior has a modest influence, it makes the predictions visibly smoother in space.

We also adapt this framework to the LSTM and to long term DMD, yet in these adaptations we have to optimise on the input initial condition rather than on the encoded initial condition. Thus, equation~\eqref{eq:AssimilationForecasting} becomes:
\begin{equation}
    \mathbf{x}_0^* = \underset{\mathbf{x}_0 \in \mathbb{R}^{N \times 2L}}{\textrm{arg min}}
    \sum_{t=0}^{T_{train}} 
    ||\mathbf{\Tilde{x}}_{t} - 
    \mathbf{x}_t(\mathbf{x}_0)||^2 + 
    \beta S(\mathbf{x}_t(\mathbf{x}_0))
\end{equation}
where $\mathbf{x}_t(\mathbf{x}_0)$ is the prediction from either the LSTM or long term DMD at time $t$ from $\mathbf{x}_0$.
For long term DMD, the computational requirements are very light because of the absence of an encoding or decoding. In contrast, for the LSTM, a time increment necessitates going through a neural network rather than performing a matrix-vector product, which is far more costly in memory and time. In particular, since using a Tikhonov spatial prior requires optimising on all pixels in parallel, it is not possible to do so for as large images with a LSTM dynamical prior as with our model as a prior.

We also try to forecast the Orléans time series in the same way with the models trained on the forest of Fontainebleau in order to test the zero-shot transfer performance of our model. 
As previously stated, the Orléans data is irregularly sampled, which makes it harder to predict with assimilation-forecasting. All assimilation-forecasting performance of the tested models with no spatial prior are reported in table~\ref{AssimilationForecastingTable}.
Using assimilation-forecasting is far more effective than a simple prediction from an initial state, no matter which model is used as the dynamical prior, as can be assessed by comparing each model's performance to those from table~\ref{ForecastingTable}.

\begin{figure}
    \centering
    \includegraphics[width=8.5cm]{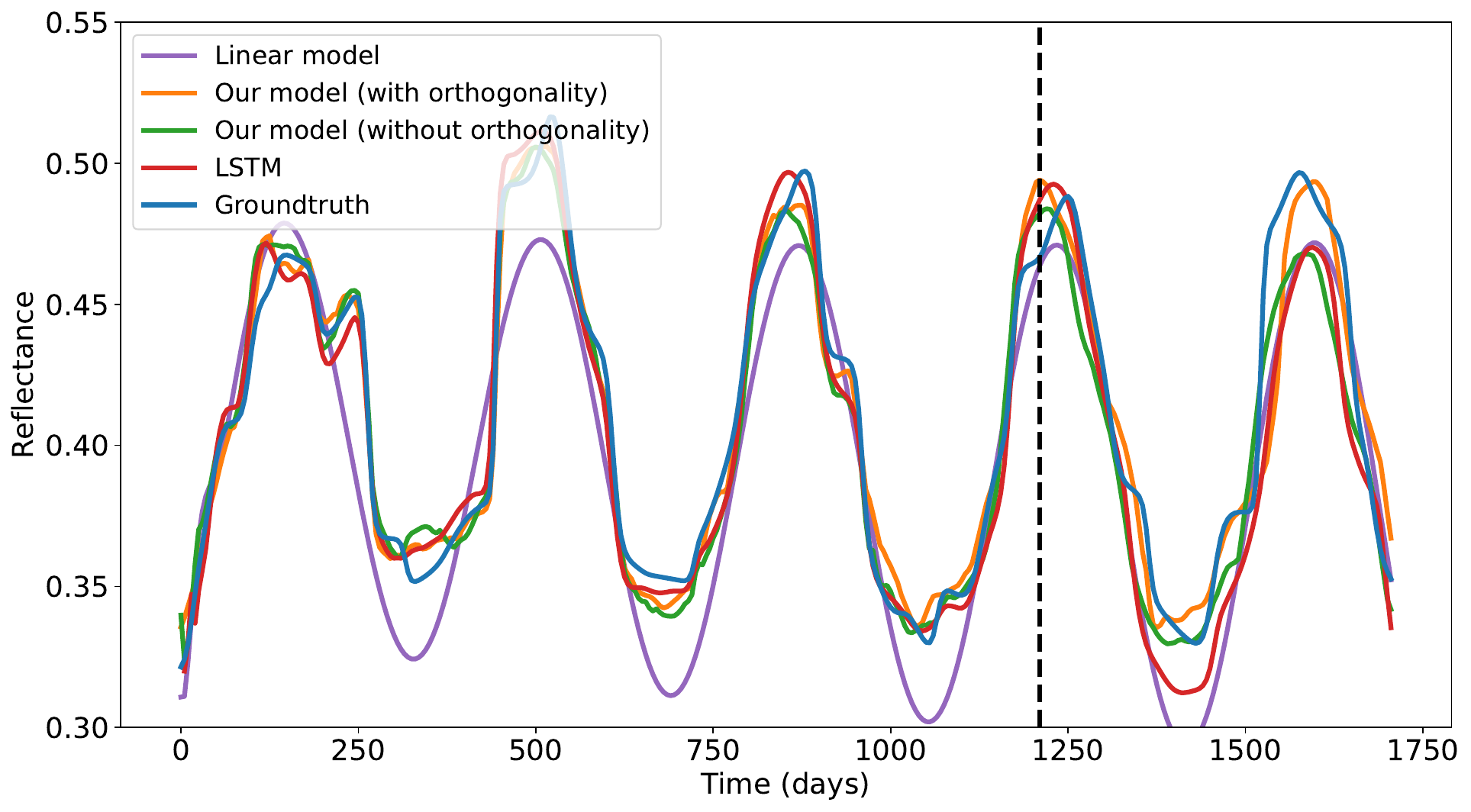}
    \caption{Assimilation-forecasting results with different dynamical priors for the reflectance of the B7 band (in near infrared). The vertical line marks the limit between the assimilated and the extrapolated data. The pixel of interest is marked by a red dot on the Fontainebleau image from figure~\ref{fig:data_samples_Sentinel}.}
    \label{fig:pred_Fontainebleau_pix}
\end{figure}

The LSTM model is more expressive than ours and it is therefore able to fit very precisely the data on which it assimilates, yet it tends to overfit on these data, which is why it performs significantly worse than ours on the extrapolation data. On the other hand, long term DMD is a very simple model, making it unable to fit the training data as well as the other models, but it performs reasonably well when transferred to the test Fontainebleau area. Our model performs the best as a dynamical prior for assimilation-forecasting, especially when trained with the orthogonality loss. Again, the gain of performance due to the orthogonality loss is higher on the test Orléans area. Qualitative results for a given pixel can be observed on figure~\ref{fig:pred_Fontainebleau_pix}, where we show the assimilation-forecasting with the best instance of each model. From this figure, one can see that  all models except for the linear one are able to fit the assimilated (training) data very closely but that the differences reside in their respective capacities to extrapolate beyond the training data.


\begin{table}[htbp]
\centering
\caption[size=9pt]{Assimilation-forecasting MSE $(\times 10^{-3})$}
\label{AssimilationForecastingTable}
\begin{tabular}{|c|c|c|}
\hline
 & Fontainebleau  & Orléans (irregular data) \\
\hline
Our model & $1.13 \pm 0.04$ & $3.46 \pm 0.20$ \\
(with orthogonality) & $(\mathbf{1.08})$ & $(\mathbf{3.22})$ \\
\hline
Our model & $1.23 \pm 0.10$ & $4.27 \pm 0.55$ \\
(no orthogonality) & $(1.13)$ & $(3.78)$ \\
\hline
LSTM & $1.50 \pm 0.08 $ & $5.42 \pm 0.84$ \\
 & $(1.38)$ & $(4.60)$ \\
\hline
Long term DMD & $2.13$ & $4.33$ \\
\hline
\end{tabular}
\end{table}

\subsection{Interpolation through data assimilation}

We now move on to performing interpolation tasks. As previously mentioned, the satellite image time series are usually incomplete since most of the observations are too cloudy to be exploited. Therefore, one often has to interpolate them in time to work with regularly sampled data. Here, we perform variational data assimilation, using our data-based model to constrain the search. Our model is trained with the procedure described in section~\ref{forecasting}, with the orthogonality loss term. The variational cost, in the framework of equation~\eqref{ConstrainedDAEquation2}, is minimised jointly on the latent initial condition and on the parameters of the pre-trained model.

\begin{figure}
    \centering
    \includegraphics[width=8.5cm]{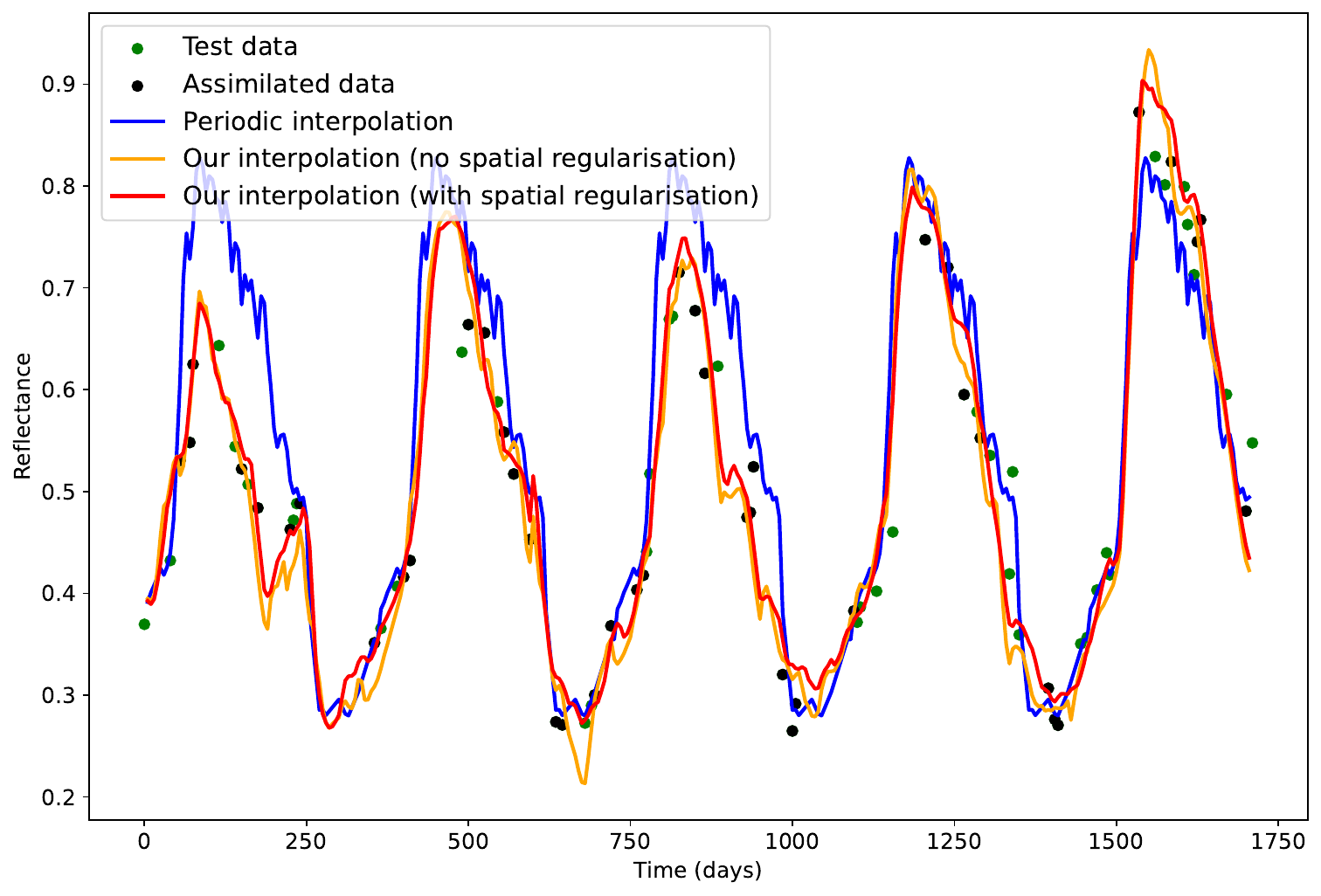}
    \caption{Comparison of interpolations for an Orléans pixel on the B7 band, using periodic interpolation and using data assimilation with our model trained on Fontainebleau data. We show an interpolation with no spatial regularisation (as can be found in~\cite{EUSIPCO}) and an interpolation with an additional spatial regularisation in the variational cost. The pixel of interest is marked by a red dot on the Orléans image on figure~\ref{fig:data_samples_Sentinel}.}
    \label{fig:interpolation_Orléans}
\end{figure}

Here, we test on raw incomplete data while our model was trained only on interpolated data from the forest of Fontainebleau. We consider the two areas in the blue squares from figure~\ref{fig:data_samples_Sentinel}.
In both cases, we have at our disposal a set of around 85 $100\times100\times10$ images, each with its associated time index, irregularly sampled over a duration of 342 time steps (i.e. nearly 5 five years). From this set, we randomly mask half of the images which we use for the interpolation, while keeping the other half as a groundtruth to evaluate the quality of the computed interpolation.
As a baseline, we seek the periodic pattern that best matches the available data, with a temporal smoothness prior using Tikhonov regularisation in the time dimension. Namely, we solve
\begin{equation}
    \underset{\mathbf{x} \in \mathbb{R}^{p \times N \times L}}{\textrm{arg min}} 
    \sum_{t \in H} ||\Tilde{\mathbf{x}}_{t} - \mathbf{x}_{t \% p}||^2 + \alpha S_t(\mathbf{x}),
\end{equation}
where $p$ is the known pseudo-period of one year, we use the notation $t\%p$ for the remainder of the Euclidean division of $t$ by $p$, and $S_t$ is the temporal smoothness prior.
This method, which we call "periodic interpolation", is a strong baseline since it explicitly leverages the physical knowledge of the pseudo-period of the data. Yet, even if it was not trained on this data but only fine-tuned on it, our model has better results, as can be seen quantitatively in table~\ref{InterpolationTable} and for a particular pixel of the forest of Orléans on figure~\ref{fig:interpolation_Orléans}. We show in the table the mean and standard deviation of the mean squared errors over 20 randomly computed masks.

Since the baseline method obtains much better results on Fontainebleau than on Orléans, the former seems to be easier to interpolate, yet the gap between this method and ours is bigger on the Orléans area since it is closer to our model's training data. 

\begin{table}[htbp]
\centering
\caption[size=9pt]{Interpolation MSE ($\times 10^{-3}$) for different areas and methods}
\label{InterpolationTable}
\begin{tabular}{|c|c|c|}
\hline
Periodic  & Fontainebleau & Orléans \\
\hline
Periodic assimilation & $0.628 \pm 0.128$ & $2.37 \pm 0.38$ \\
\hline
Our method & $0.349 \pm 0.154$ & $2.22 \pm 0.37$ \\
\hline
Our method & $\mathbf{0.335 \pm 0.146}$ & $\mathbf{2.13 \pm 0.34}$ \\
(with spatial prior) & &  \\
\hline
\end{tabular}

\end{table}

\subsection{Training on an irregular time series}

As our final experiment, we investigate the training of our architecture on an irregular version of the Fontainebleau time series. This corresponds to the relatively simple setting mentioned in subsection~\ref{Irregular} since this time series results from a regular sampling from which some observations have been removed because they were not usable. We were therefore able to optimise directly on the discrete operator $\mathbf{K}$ rather than on its continuous counterpart $\mathbf{L}$. As explained in subsection~\ref{Irregular}, one just has to adapt the prediction, auto-encoding and linearity loss terms by computing them only for time delays for which the groundtruth is available. We were able to obtain satisfying results in this way, although the computed model is not as good as when training on interpolated time series. This tends to suggest that training our model directly on irregular time series can be a possibility when it is not possible to perform an interpolation as a pre-processing step.

We found that training on irregular data makes our model more subject to overfitting. Indeed, the model is not forced to predict a smooth evolution anymore but only to be able to correctly reconstruct some sparsely located points. Therefore, all regularisation terms that we presented in subsection~\ref{Training} are very important to get the most out of this dataset.  To support this claim, we performed an ablation study in which we tested different loss functions: the complete loss with the 4 terms from equations~\eqref{pred}-\eqref{orth} and 4 versions where one of the terms has been removed. For each version of the loss function, we trained models on the Fontainebleau data from 5 different initialisations. We then retrieved the mean and standard deviations of the mean squared errors obtained when performing assimilation-forecasting as in subsection~\ref{AssimilationForecasting}. The results are presented in table~\ref{AblationStudy}. 
One can see that the final results on the Fontainebleau area largely depend on the model initialisation, yet both the mean and the standard deviation of the MSE are lower when using all loss terms.

\begin{table}[htbp]
\centering
\caption[size=9pt]{Forecasting MSE ($\times 10^{-3}$) of models trained on irregularly-sampled data with different loss functions}
\label{AblationStudy}

\begin{tabular}{|c|c|c|c|}
\hline
 & Fontainebleau  & Orléans \\
\hline
Complete loss & $\mathbf{0.699 \pm 0.130}$ & $\mathbf{3.480 \pm 0.198}$ \\
\hline
No orthogonality & $0.922 \pm 0.233$ & $3.564 \pm 0.301$ \\
\hline
No linearity & $2.722 \pm 0.576$ &  $5.454 \pm 0.515$ \\
\hline
No auto-encoding & $1.252 \pm 0.128$ & $3.770 \pm 0.172$ \\
\hline
No prediction & $3.514 \pm 1.276$ & $4.586 \pm 0.232$ \\
\hline
\end{tabular}

\end{table}

\begin{figure}
    \centering
    \includegraphics[width=8.5cm]{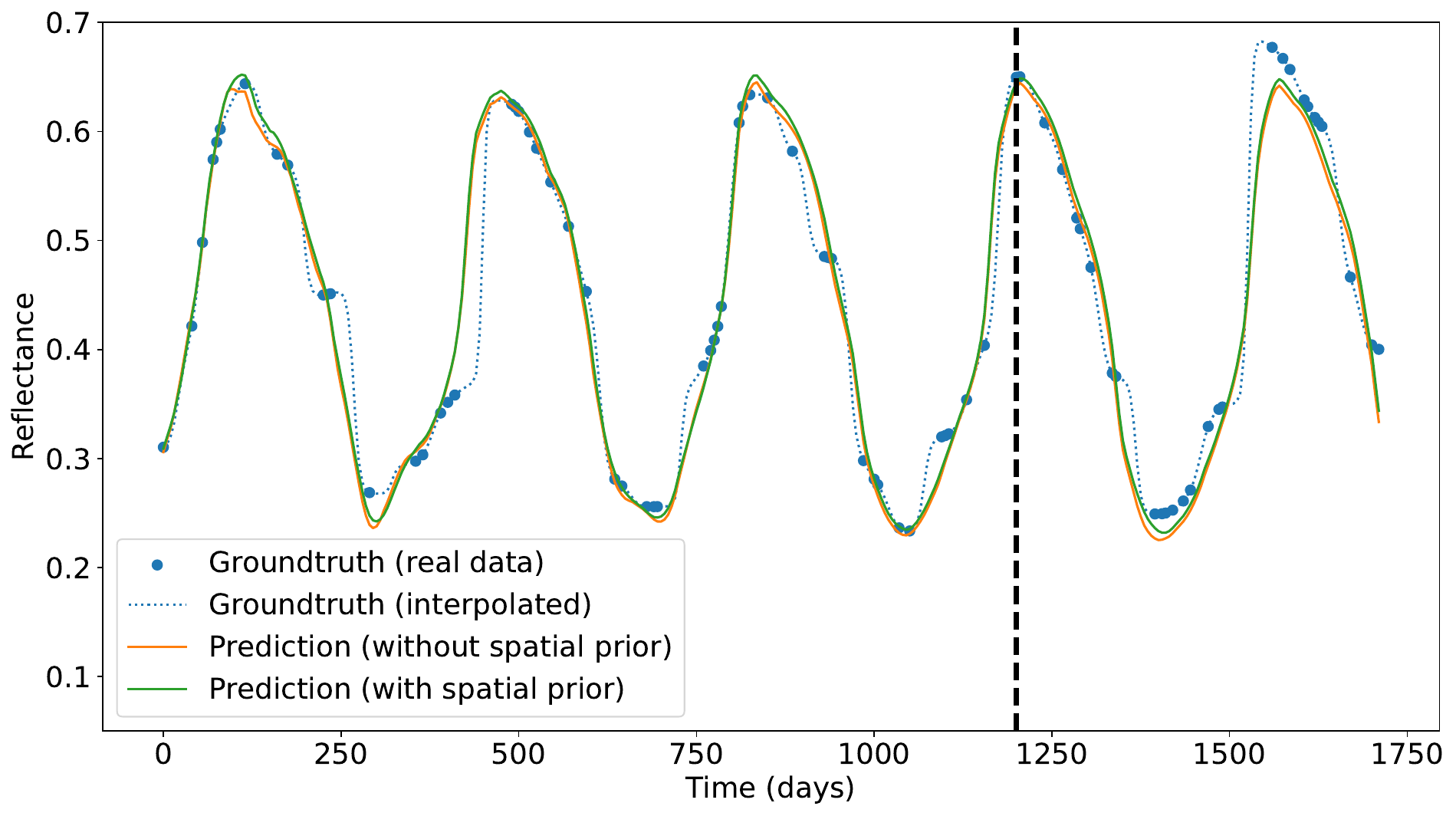}
    \caption{Forecasting results on the B7 band, with irregular data from the forest of Fontainebleau. The considered pixel is marked by a blue dot on figure~\ref{fig:data_samples_Sentinel}.}
    \label{fig:Irregular_assimilation_forecasting}
\end{figure}

We show qualitative results of the assimilation-forecasting of irregular data using one of our models trained on irregular data with the complete loss function in figure~\ref{fig:Irregular_assimilation_forecasting}. We emphasize that the blue curve is only a Cressman interpolation of the groundtruth points and should not be seen as a groundtruth here. Our model fits the training points well and, in some way, performs a smoother interpolation than the Cressman method that was used to obtain the regularly-sampled data.

We emphasize that, when tested on the same irregular test data, models trained on interpolated Fontainebleau data have better interpolation performance but lower forecasting performance than models trained on irregular Fontainebleau data. Using interpolation as a pre-processing step is not a trivial choice since models trained on these data will learn the interpolation scheme along with the true data. However, it can be seen as a form of data augmentation.

\section{Discussion}

In the assimilation results presented in the previous section, adding a regularisation on the spatial gradient generally only results in a modest improvement of the mean squared error compared to using no regularisation. Since the chosen regularisation was a very basic one, this suggests that our predictions could gain more from spatial information. One could imagine using a more complex spatial prior, e.g. a data-driven one like in~\cite{4DVarNet}. One could also use an image as the input of the Koopman auto-encoder instead of a single pixel, which would enable it to directly leverage spatial information (e.g. with a convolutional architecture) but might be more difficult and less suited to pixel-level downstream tasks. Another possibility is to train a convolutional neural network to correct the residual errors made when recomposing images from our pixelwise model. This approach has been presented in~\cite{EUSIPCO} and it can indeed improve the results. Yet, it is much less flexible and elegant since such a CNN needs to be trained for every new model or task considered, for which we do not necessarily have enough time or data.

The weights of the prior terms in assimilation should be chosen carefully, although a slightly inaccurate choice is unlikely to severely affect the results. A good way to proceed, when possible, is to use a set of validation data to set the parameters $\alpha$ and $\beta$ from equations~\eqref{classicalDAequation}-\eqref{ConstrainedDAEquation2}. About the choice of allowing the parameters of the pretrained model to vary when performing data assimilation (i.e. solving the problem from equation~\eqref{ConstrainedDAequation} or~\eqref{ConstrainedDAEquation2}), it seems from our experiments that:
(1) It is more beneficial to make the parameters vary when working with data that differ from what could be found in the training dataset. This can be seen as fine-tuning the model.
(2) Allowing the model parameters to vary is effective when performing interpolation, but more dangerous for forecasting. An explanation could be that a slightly modified model keeps its tendency to generate smooth trajectories but not necessarily its long-term stability. One could investigate extensions of the framework of equation~\eqref{ConstrainedDAEquation2} with e.g. an orthogonality term to make sure that the modified model remains stable.
\section{Conclusion}
In this paper, we presented a method that enables to jointly learn a Koopman invariant subspace and an associated Koopman matrix of a dynamical system in a data-driven way. We showed that this method enables to learn a continuous representation of dynamical systems from discrete data, even in difficult contexts where the data are sparsely or irregularly sampled. In addition, it was demonstrated that a trained model is not only useful to forecast the future state of a dynamical system but also to solve downstream tasks.  Indeed, we used the forward prediction as a pretext task to learn general useful information about the dynamical system in a self-supervised way. Since our architecture is fully differentiable, we showed how this information can be leveraged to solve inverse problems using variational data assimilation.

A possible extension of our work is to introduce a control variable in order to better predict the state of systems on which we know that some information is lacking. For example, precipitation data could be used as a control variable to better predict the vegetation reflectance. For image data specifically, one could make a better use of the spatial structure of the images by learning a complex spatial prior that would be coupled to the dynamical prior or by directly learning an end-to-end model that takes into account both dynamical and spatial information. Finally, a stochastic extension of our framework would make it able to output distributions of possible trajectories rather than single predictions.



\appendices

\section{Proof of Theorem~\ref{thm:periodic}}
\label{orthogonality}

Let us consider linear dynamics in a latent space given by $\mathbf{z}_t \in \mathbb{R}^{d}$, and
\begin{equation}
    \mathbf{z}_{t+1} = \mathbf{Kz}_t
\end{equation}
where $\mathbf{K} \in \mathcal{SO}(d)$. $\mathbf{K}$ belongs to the special orthogonal group, i.e. the group of invertible matrices satisfying $\mathbf{KK}^T = \mathbf{K}^T\mathbf{K} =  \mathbf{I}$, and with determinant equal to $+1$.
First we note that the norm of the iterates $\mathbf{z}_t$ remain equal to that of the initial condition $\mathbf{z}_0$. Indeed:
\begin{equation}
    ||\mathbf{z}_{t+1}||^2 = ||\mathbf{Kz}_{t}||^2 = \mathbf{z}_t^T \mathbf{K}^T\mathbf{K} \mathbf{z}_t = \mathbf{z}_t^T\mathbf{z}_t = ||\mathbf{z}_t||^2
\end{equation}
and it is easy to see by induction that every iterate's norm is equal to $||\mathbf{z}_0||$. So the dynamics remain on a sphere of radius $||\mathbf{z}_0||$. This is equivalent to saying that the matrix group $\mathcal{SO}(d)$ acts on the sphere via matrix-vector multiplication.

Besides, $\mathcal{SO}(d)$ is a Lie group, whose Lie algebra $\mathfrak{so}(d)$ is the set of skew-symmetric matrices of size $d$. Furthermore, as $\mathcal{SO}(d)$ is compact, the exponential map $\exp : \mathfrak{so}(d) \to \mathcal{SO}(d)$, corresponding here to the matrix exponential, is surjective~\cite{brocker2013representations}. This means that any special orthogonal matrix can be written as the matrix exponential of a skew-symmetric matrix $\mathbf{L}$: $\exp(\mathbf{L}) = \mathbf{K}$. Equivalently, a skew-symmetric matrix logarithm of a special orthogonal matrix always exists. In these conditions, $\mathbf{z}_{t+1}$ is the solution to the following ODE, representing the same dynamics in continuous time:
\begin{equation}
    \frac{d\mathbf{z}}{dt} = \mathbf{Lz}
\end{equation}
with $\mathbf{z}(0) = \mathbf{z}_t$.
We proceed to show that the dynamics generated by this ODE must be periodic.
$\mathbf{L}$ is a skew-symmetric matrix, to which the spectral theorem applies: it can be diagonalized in a unitary basis, and its eigenvalues must be purely imaginary. 
Denoting $\mathcal{U}(d)$ the set of unitary matrices of size $d$, there exists $\mathbf{U} \in \mathcal{U}(d)$ such that:
\begin{equation}
    \mathbf{L} = \mathbf{U}^* \mathbf{D} \mathbf{U}
\end{equation}
with $\mathbf{D} = \textrm{diag}(i \alpha_1,i \alpha_2,...,i \alpha_d), \alpha_k \in \mathbb{R}.$ By denoting $\mathbf{K}^\tau = \exp(\tau\mathbf{L})$ (giving $\mathbf{z}_t$ by matrix multiplication with $\mathbf{z}_0$), we can write:
\begin{equation}
\mathbf{K}^\tau = \exp(\tau\mathbf{L}) = \mathbf{U}^* \exp(\tau\mathbf{D}) \mathbf{U}
\end{equation}
If we write out $\mathbf{K}^\tau_{rs}$, denoting as $\mathbf{u}_r$ the $r\textsuperscript{th}$ column of $\mathbf{U}$, we get 
\begin{equation}
    \mathbf{K}^\tau_{rs} = \mathbf{u}_r^* \exp(\tau \mathbf{D}) \mathbf{u}_s = \sum_{k=1}^{d} u_{kr}^* \exp(i\tau \alpha_k) u_{sk}.
\end{equation}
The exponential factors are periodic with periods $\frac{2\pi}{\alpha_k}$. Hence each entry of $\mathbf{K}^\tau$ is a linear combination of periodic functions. Mathematically, such a linear combination is only periodic when all the ratios between pairs of periods of the summands are rational. For all practical purposes, however, when numbers are represented with finite precision in a computer, such a linear combination can be itself seen as periodic. Finally, the same argument applies for all entries, with a common period, so the whole matrix $\mathbf{K}^\tau$ is periodic.
\bibliographystyle{IEEEtran}
\bibliography{references.bib}

\end{document}